\documentclass[journal]{IEEEtran}
\ifCLASSINFOpdf
  \usepackage[pdftex]{graphicx}
\else
\fi
\usepackage[cmex10]{amsmath}
\usepackage{amsfonts}
\usepackage{amssymb}\usepackage{amsthm}\usepackage{bm}

\usepackage{amsfonts}

\usepackage{algorithmic}
\usepackage{algorithm}
\usepackage{boxedminipage}

\usepackage{xcolor}

\usepackage[normalem]{ulem}
\usepackage{verbatim}

\begin{document}
%
\title{Neural Networks with Activation Networks}
%
%
%

\author{Jinhyeok~Jang,
	Jaehong~Kim,
	Jaeyeon~Lee,
        and~Seungjoon~Yang
\thanks{J. Jang, J. Kim, and J. Lee are with Electronics and Telecommunications Research Institute (ETRI), Daejeon, Korea (e-mail:jangjh6297@gmail.com, jhkim504@etri.re.kr, leejy@etri.re.kr).}
\thanks{
S. Yang is with School of Electrical and Computer Engineering, Ulsan National Institute of Science and Technology
(UNIST), Ulsan, Korea (e-mail: syang@unist.ac.kr). 
}
\thanks{
This work was supported by the ICT R\&D program of Ministry of Science and ICT / Institute for Information \& Communications Technology Promotion under Grant (2017-0-00162, Development of Human-care Robot Technology for Aging Society)}
\thanks{
This work was supported by the Ulsan National Institute of Science and Technology Free Innovation Research Fund under Grant 1.170067.01
}
}

\maketitle

\begin{abstract}
This work presents an adaptive activation method for neural networks that exploits the interdependency of features. Each pixel, node, and layer is assigned with a polynomial activation function, whose coefficients are provided by an auxiliary activation network. The activation of a feature depends on the features of neighboring pixels in a convolutional layer and other nodes in a dense layer. The dependency is learned from data by the activation networks. In our experiments, networks with activation networks provide significant performance improvement compared to the baseline networks on which they are built. The proposed method can be used to improve the network performance as an alternative to increasing the number of nodes and layers.
\end{abstract}

\ifCLASSOPTIONpeerreview
\begin{center} \bfseries EDICS Category: 3-BBND \end{center}
\fi
%
\IEEEpeerreviewmaketitle

\section{Introduction}
%
%
%
%

\IEEEPARstart{N}{eural} networks often use a single function as an activation function for all nodes in the networks. The activation of a node output only depends on the intermediate output of the node computed by a dense or convolutional layer. The shape of an activation function determines how the node output is triggered. In general, strong features extracted by a dense or convolutional layer are passed through to the next layer by an activation function. Neural networks with a single activation function have been successfully applied to various problems. However, some studies have used more complicated activation methods to improve the network performance \cite{dushkoff2016adaptive, harmon2017activation, piazza1992artificial, chung2016deep}.  

The activation of features in a biological system can be affected by many factors. For example, features that have similar characteristics with strong energy can inhibit the activation of a certain feature \cite{stromeyer1972spatial}. Features in a sensory input can be affected by the sensors at neighboring locations \cite{bridgeman1971metacontrast, macknik2004spatial}. Overt and covert attention can encourage or discourage the activation in certain regions \cite{kuhn2009you, wolfe2004attributes}. Neural networks that mimic the activation behavior of biological systems have been studied. In \cite{amari1977dynamics, pinto2001spatially, fernandes2013lateral}, the lateral inhibition model is introduced to neural networks. In \cite{amari1977dynamics, pinto2001spatially}, the effect of the excitation of features at spatially coupled locations on the excitation of a feature in time is studied. By assuming that lateral inhibition instantaneously occurs, simpler lateral inhibition networks without time dependency can be considered. In \cite{fernandes2013lateral}, a box filter is applied to the intermediate outputs of a convolutional layer. The activation at the center pixel location in a convolutional layer is inhibited by strong intermediate outputs of neighboring pixels computed by a box filter. In \cite{itti1998model, xuk2015show}, the attention model is introduced to recurrent networks. In \cite{itti1998model}, the saliency of features are found in an input image, and a neural network is designed to weigh the features based on the saliency. In \cite{xuk2015show}, the activation of a node in a recurrent network is modulated by a variable representing attention. The recurrent network shifts its attention to certain regions as it makes decisions. 

This work presents an activation method for neural networks that exploits the interdependency of features. In particular, the shapes of the activation functions for each pixel, node, and layer and the dependency of the activation on other features are learned by an auxiliary activation network. A polynomial function is used as an activation function, whose coefficients are determined by an activation network. Each layer in a deep network is accommodated with the activation network. The network learns to adaptively use different activations for each pixel, node, and layer for a given input depending on the features of the neighboring pixels in a convolutional layer and other nodes in a dense layer. Unlike the lateral inhibition and attention models, where the interdependency of features is predefined based on the domain knowledge, the proposed method learns the dependency from the data. 

The proposed activation networks are applied to deep convolutional neural networks (CNNs). We prepared deep networks for object recognition and denoising for analysis. The addition of activation networks increases the number of parameters in a network. However, the networks with activation networks were trained faster with lower training losses. The networks with activation networks outperformed both the baseline networks on which they were built and deeper and wider versions of the networks with higher computational complexity. The node outputs of the proposed networks show that the activation networks enable a network to utilize high-level features even from the first layers. This aspect contrasts the usual operations of deep networks, where deeper layers utilize high-level features that are built up from low-level features of shallower layers. We also compared the performance of the proposed networks to various activation methods. In our experiments, providing the adaptive activation that exploits the interdependency of features with activation networks significantly improved the network performance at modest increases of computational complexity. The proposed method can be used to improve the network performance as an alternative to increasing the number of nodes and layers.

The remainder of this paper is organized as follows. The use of activation networks is introduced in Section \ref{sec:proposed}. The proposed method is compared to various adaptive activation methods in Section \ref{sec:comparison}. The performance of the proposed networks is evaluated with deep CNNs for object recognition and denoising in Section \ref{sec:ex}. Section \ref{sec:conclusion} concludes the paper.

\section{Neural Networks with Activation Networks}

\subsection{Activation Networks with Polynomial Activation Functions}
\label{sec:proposed}

\begin{figure}[h]
	\centering
	\begin{minipage}{0.5\linewidth}
		\centering
		{\includegraphics[trim=150 150 180 150, clip, width=\linewidth]{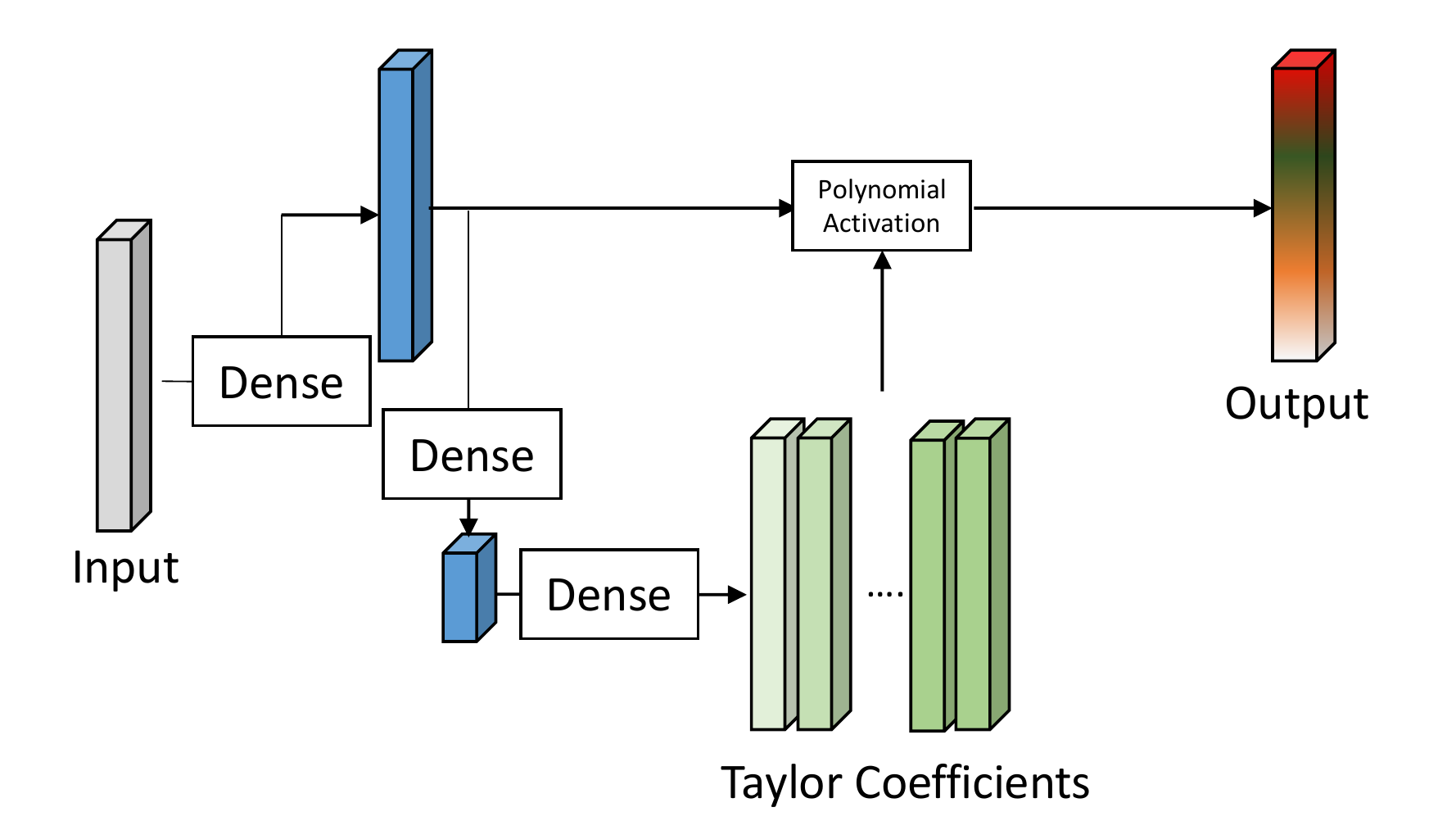}}%

		{\footnotesize (a)}
	\end{minipage}%
	\begin{minipage}{0.5\linewidth}
		\centering
		{\includegraphics[trim=150 150 180 150, clip, width=\linewidth]{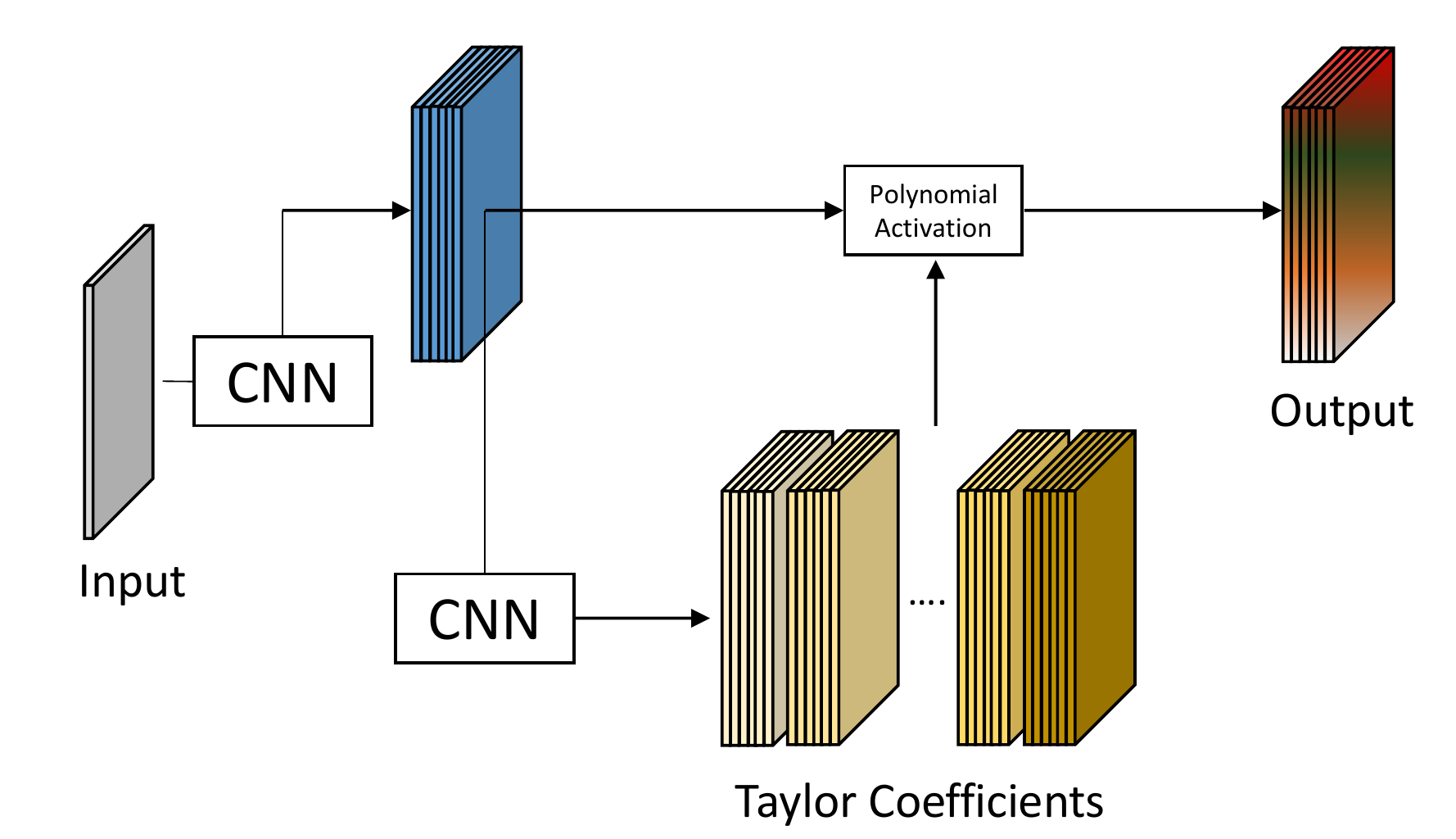}}%
		
		{\footnotesize (b)}
	\end{minipage}%
	
	\caption{Layer schematics of a network with the activation network:
		(a) dense layer;
		(b) convolutional layer.
		}
	\label{fig:schematics}
\end{figure}

This work considers a general form of activation, where a network learns to use different shapes of activation functions for the pixels, nodes, and layers depending on other features of neighboring pixels in a convolutional layer and other nodes in a dense layer. Consider a network whose operations of the $l$th layer are given as follows. For a dense layer, an intermediate output of the $i$th node $u^l_i$ is obtained by
\begin{equation}
	u^l_i = \sum_j^{n_{l-1}} W^l_{ij} x^{l-1}_j
\end{equation}
for $i=1, 2, \cdots, n_l$, where $x^{l-1}_j$ is the output of the $j$th node in the $(l-1)$th layer, and $W^l_{ij}$ is the weight. A set of polynomial coefficients is obtained by an auxiliary network:
\begin{equation}
	a^l_{ki} = \sum_{j=1}^{n_l} V^l_{kj} u^l_j + b^l_{ki}
	\label{eq:actnetdense}
\end{equation}
for $k=0,1,\cdots, K$, where $V^l_{ij}$ and $b^l_i$ are the weights and a bias for a node that provides the $k$th-order coefficient of the $i$th node $a^l_{ki}$, respectively. The intermediate output is activated by a $K$th-order polynomial function as follows:
\begin{equation}
	x^l_i = \sum_{k=0}^K a^l_{ki} (u^l_i)^k.
	\label{eq:actdense}
\end{equation}
For a convolutional layer, an intermediate output of the $i$th node $u_i^l$ is obtained by
\begin{equation}
	u_i^l = \hbox{conv}(w_i^l, x^{l-1})
\end{equation}
for $i=1, 2, \cdots, n_1$, where $w_i^l$ is the filters, and $x^{l-1}$ is the set of node outputs of the $(l-1)$th layer. A set of polynomial coefficients is obtained by 
\begin{equation}
	a^l_{ki} = \hbox{conv}(v^l_{ki}, b^l_{ki}, u^l)
	\label{eq:actnetconv}
\end{equation}
for $k=0,1,\cdots, K$, where $v^l_{ki}$ and $b^l_{ki}$ are a set of filters and bias for a node that provides the $k$th-order coefficients $a^l_{ki}$, respectively. The intermediate output is activated by
\begin{equation}
	x^l_i(\xi, \zeta) = \sum_{k=0}^K a^l_{ki}(\xi, \zeta) (u^l_i(\xi, \zeta))^k,
	\label{eq:actconv}
\end{equation}
where $(\xi, \zeta)$ are pixel indices. The layer schematics of the proposed architecture are shown in Fig. \ref{fig:schematics}

The activation function for the $i$th node of a dense layer and the $(\xi, \zeta)$th pixel of a convolutional layer is the $K$th-order polynomial function. The shapes of the activation functions are determined by the coefficients provided by the auxiliary dense and convolutional networks in \eqref{eq:actnetdense} and \eqref{eq:actnetconv}, respectively. We name the auxiliary network an activation network. The activation network provides different shapes of polynomial activation functions for each pixel, node, and layer. The activation depends on the features of other nodes in a dense layer and neighboring pixels in a convolutional layer. The dependency is learned by the activation network from the data during the training.

The activation networks are simultaneously trained with the original network. The weights of the original network are updated by
\begin{equation}
	W^l_{ij} \gets W^l_{ij} - \eta \frac{\partial E}{\partial W^l_{ij}} \quad \hbox{or} \quad w^l_{i} \gets w^l_{i} - \eta \frac{\partial E}{\partial w^l_{i}} 
\end{equation}
for a dense or convolutional layer, where $E$ is the cost function for the training. The weights and biases of the activation networks are simultaneously updated by
\begin{equation}
	V^l_{ki} \gets V^l_{ki} - \eta \frac{\partial E}{\partial V^l_{ki}} \quad \hbox{or} \quad v^l_{ki} \gets v^l_{ki} - \eta \frac{\partial E}{\partial v^l_{ki}}
\end{equation}
for a dense or convolutional activation network, resepectively, and 
\begin{equation}
	b^l_{i} \gets b^l_{i} - \eta \frac{\partial E}{\partial b^l_{i}}
\end{equation}
using the same cost function $E$.

The activation networks includes the use of usual activation functions. If $V^l_{kj}$'s or $v^l_{ki}$'s are set to zero for all $k$ and $j$, and $b^l_{ki}$'s are set to predefined constants in \eqref{eq:actnetdense} or \eqref{eq:actnetconv}, one can use a well-known activation function. For example, $b^l_{ki}$'s can be set to approximate the sigmoid or hyperbolic tangent functions, whose Talylor expansions are given by
\begin{equation}
	\begin{split}
	\mathrm{sigmoid}(u) & = \frac{1}{2} + \frac{1}{4}u-\frac{1}{48}u^3+\frac{1}{480}u^5+o(u^6) \\
	\tanh(u) & = u - \frac{1}{3}u^3 + \frac{2}{15}u^5 - \frac{17}{315} u^7+o(u^8).
	\end{split}
\end{equation}
If $b^l_{ki}$'s are set to the same constants for all nodes, one can use a single activation function in a network. Hence, the network with activation networks includes networks that use a single activation function.

\subsection{Comparisons to Other Activation Methods}
\label{sec:comparison}

Even though neural networks with the same activation function for all nodes have been used successfully for many applications, some studies have used variations of the activation functions to improve the performance. In \cite{dushkoff2016adaptive, harmon2017activation}, multiple activation functions, $\phi_k$'s, are applied to the intermediate output of a node, and their linear combination is used as an activation, i.e.,
\begin{equation}
	x^l_i = \sum_k \alpha^l_k \phi_k(u^l_i).
\end{equation}
The coefficients for the linear combination, $\alpha^l_k$'s, are learned from the data during the training of a network. The network uses a different activation function for each layer. In \cite{gulcehre2016noisy}, a sigmoid activation function is fed with a noisy intermediate output,
\begin{equation}
	x^l_i = \phi(u^l_i + n)
\end{equation}
where $n$ is the noise. The result is to randomly change the slope of the activation function for a given input. Hence, it has an effect of using a different activation function for each node. Compared to these approaches, the proposed method uses more general shapes of activation functions, and the variations of the activation functions are not limited to being node-wise and can be extended to being pixel-wise. 

In \cite{piazza1992artificial, chung2016deep}, a polynomial function
\begin{equation}
	x^l_i = \sum_{k=0}^K a^l_{k} (u^l_i)^k
\end{equation}
 is used as an activation function for the node-wise variant activation. Once the shapes of the polynomial activation functions are learned during the training, the shapes remain fixed during the usage of the network. In contrast, we train the activation networks that provide different shapes of activation functions adaptively when the input changes.  

In \cite{fernandes2013lateral}, the lateral inhibition by neighboring nodes is simulated by shifting the bias of an activation function by
\begin{equation}
	x^l_i = \phi(u^l_i-\iota^l_i).
	\label{eq:inhibition}
\end{equation}
The quantity $\iota^l_i$ is the output of the so-called receptor field given by
\begin{equation}
	\iota^l_i \propto \sum_{j\in {\cal N} (i)} u^l_j - u^l_i,
\end{equation}
where $u^l$'s are the intermediate outputs of the nodes and ${\cal N}(i)$ is the set of neighboring nodes of the $i$th node. In this approach, how other node outputs affect the activation of a node is predetermined by the design of the activation function and receptive fields. In contrast, we allow a network to learn how other node outputs should affect the activation of a node from data during training. 

In \cite{xuk2015show}, the attention model is used to enable a recurrent network to pay attention to particular features as time proceeds. The activation function in the last layer is modulated by the attention, or
\begin{equation}
	x^l_i = \alpha(u^l_i) \phi(u^l_i),
	\label{eq:attention}
\end{equation}
where $\alpha(u^l_i)$ represents the attention of a network. The attention is applied in the last layer that makes a final decision. The recurrent network can rely more heavily on some features when the attention shifts in time. In contrast, with the proposed method, the variation of activation is not limited to the modulation but learned for a specific task from the data.

\section{Experiments}
\label{sec:ex}

\subsection{LeNet for Object Recognition}
\label{sec:lenet}

\begin{figure}[h]
	\centering
	\begin{minipage}{0.4\linewidth}
		\centering
		{\includegraphics[width=\linewidth]{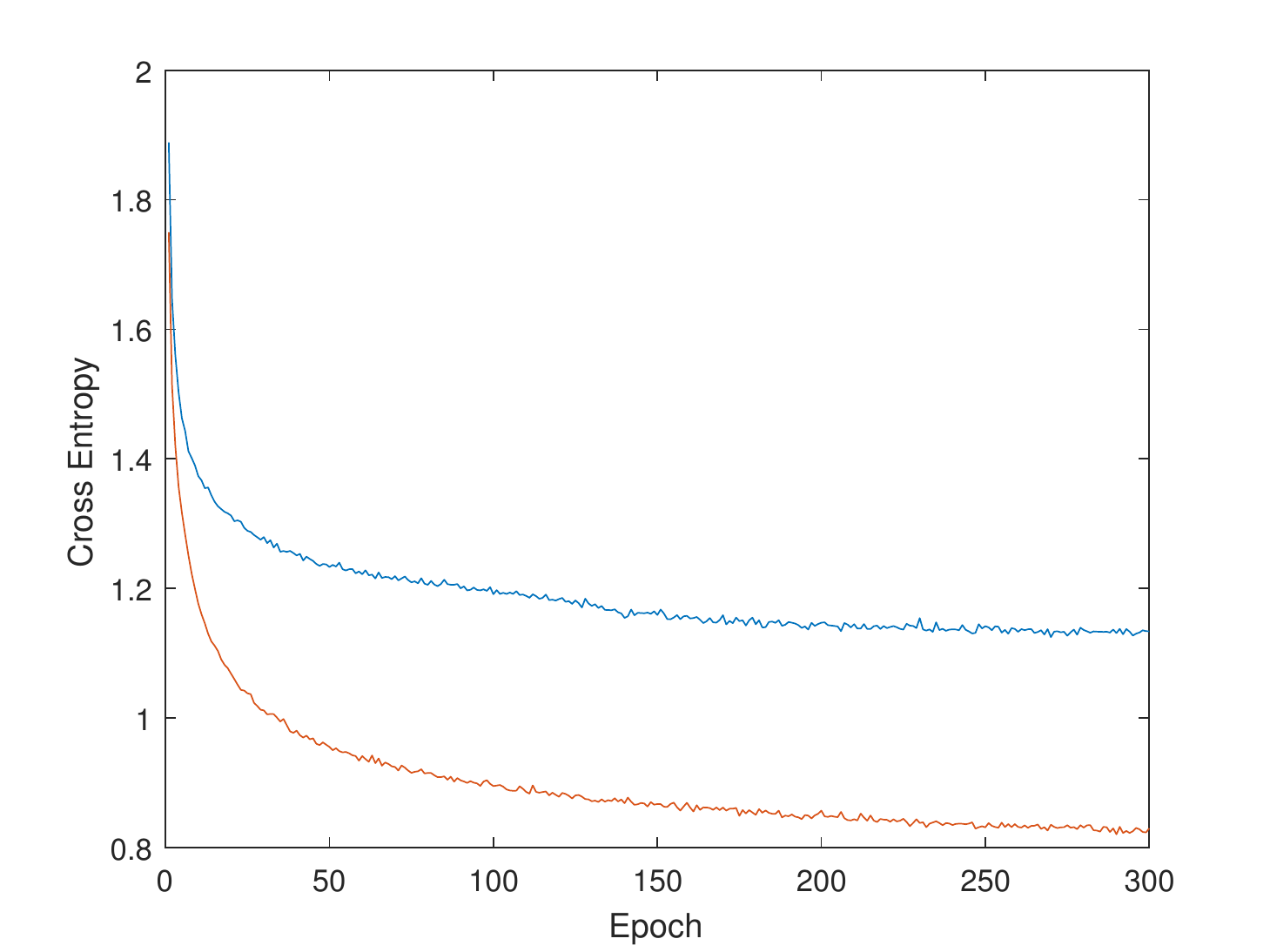}}%

		{\footnotesize (a)}
	\end{minipage}%
	\begin{minipage}{0.4\linewidth}
		\centering
		{\includegraphics[width=\linewidth]{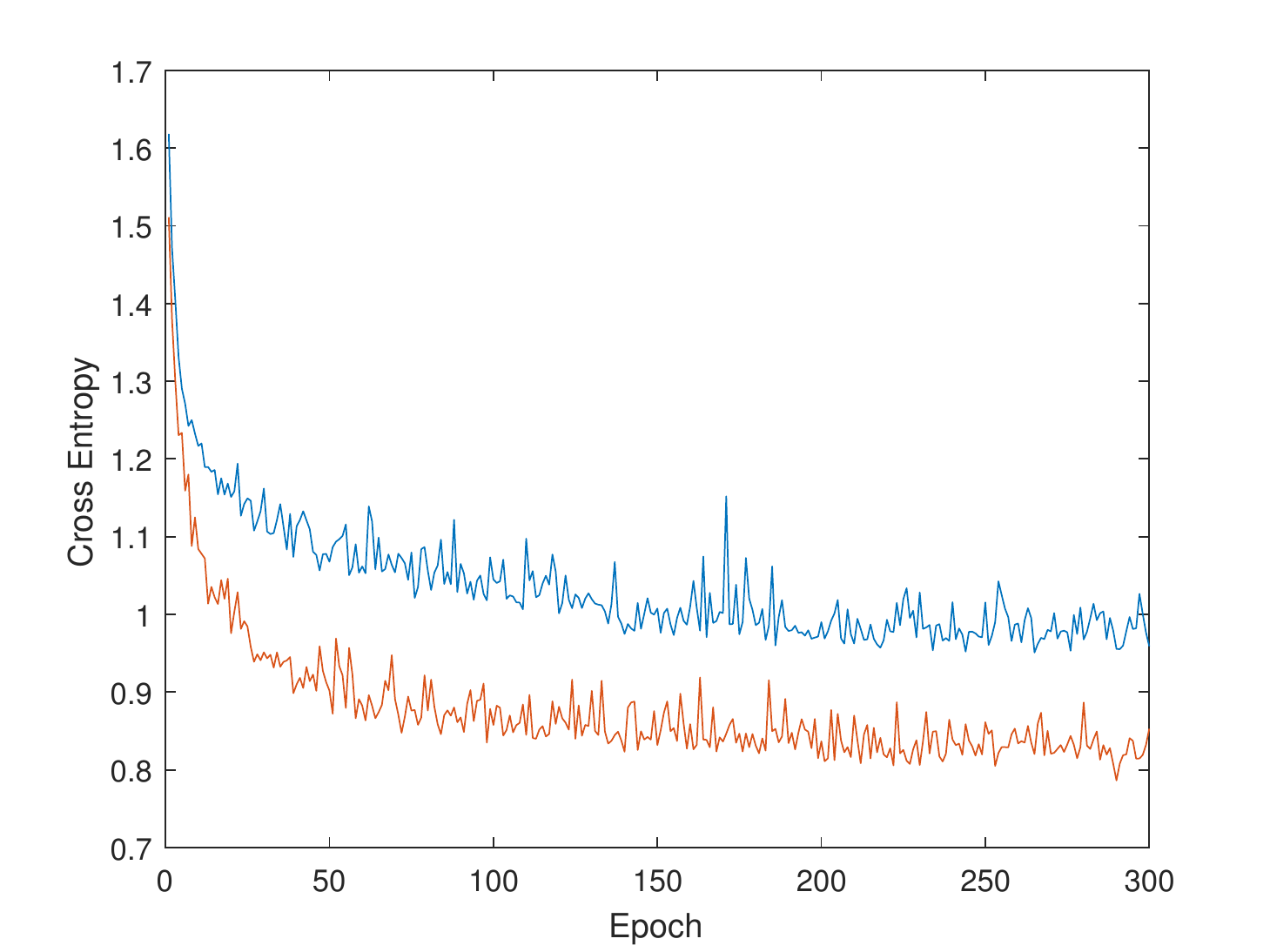}}%
		
		{\footnotesize (b)}
	\end{minipage}%
	
	\caption{Training of LeNet with CIFAR-10:
		(a) training loss;
		(b) validation loss.
		blue: baseline LeNet;
		red: LeNet-AN.
		}
	\label{fig:lenettraining}
\end{figure}

To analyze the deep networks with activation networks, we prepare LeNet \cite{krizhevsky2012imagenet} by adding the proposed activation network for each convolutional and dense layer except for the last layer, which is prepared for object recognition. The order of the polynomial activation function $K$ is determined experimentally and set to 5. The LeNet with the activation network is denoted by LeNet-AN. A baseline LeNet with the ReLU activation function is prepared for comparison. The networks are implemented with Keras deep learning package \cite{chollet2015keras} and trained with the CIFAR-10 dataset \cite{krizhevsky2009learning}. To rule out other factors on the performance, no data augmentation is used for training. Fig. \ref{fig:lenettraining} shows the training and validation losses for the baseline LeNet and LeNet-AN. LeNet-AN was trained much faster with smaller training and validation losses. 

\begin{figure}[h]
	\centering
	\begin{minipage}{0.3\linewidth}
		\centering
		{\includegraphics[trim=64 49 65 23, clip, width=0.5\linewidth]{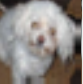}}%
		{\includegraphics[trim=64 49 65 23, clip, width=0.5\linewidth]{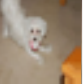}}
		
		{\footnotesize (a)}
	\end{minipage}
	\begin{minipage}{0.3\linewidth}
		\centering
		{\includegraphics[trim=64 49 65 23, clip, width=0.5\linewidth]{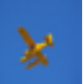}}%
		{\includegraphics[trim=64 49 65 23, clip, width=0.5\linewidth]{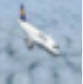}}
		
		{\footnotesize (b)}
	\end{minipage}%

	\caption{Two input images for Figs. \ref{fig:lenetact} and  \ref{fig:lenetnodeactivation}:
	(a) dog 1, 2;
	(b) airplane 1, 2.
	}
	\label{fig:lenetimages}
\end{figure}

\begin{figure}[h]
	\centering
	\begin{minipage}{\linewidth}
		\centering
		{\includegraphics[width=0.33\linewidth]{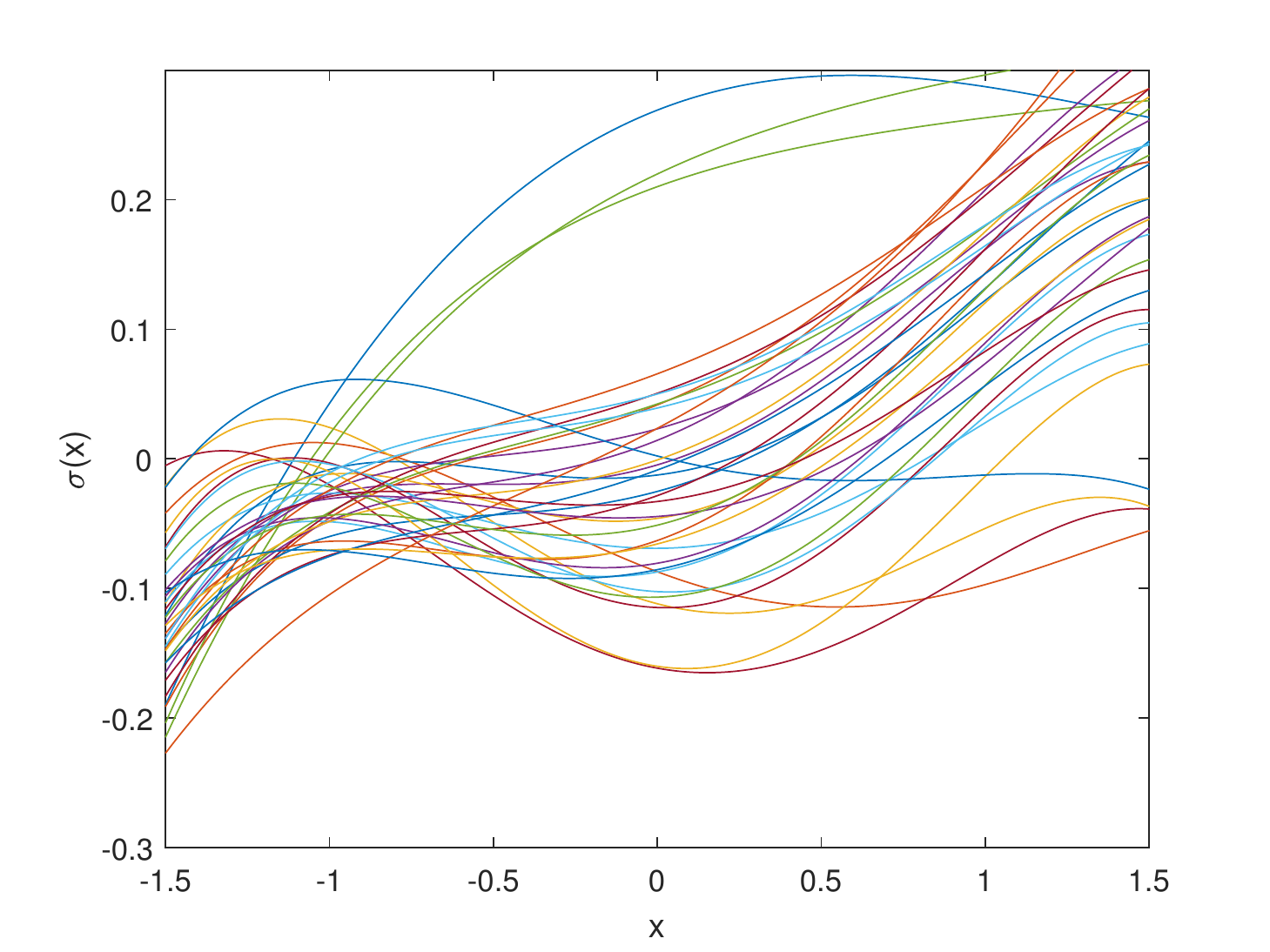}}%
		{\includegraphics[width=0.33\linewidth]{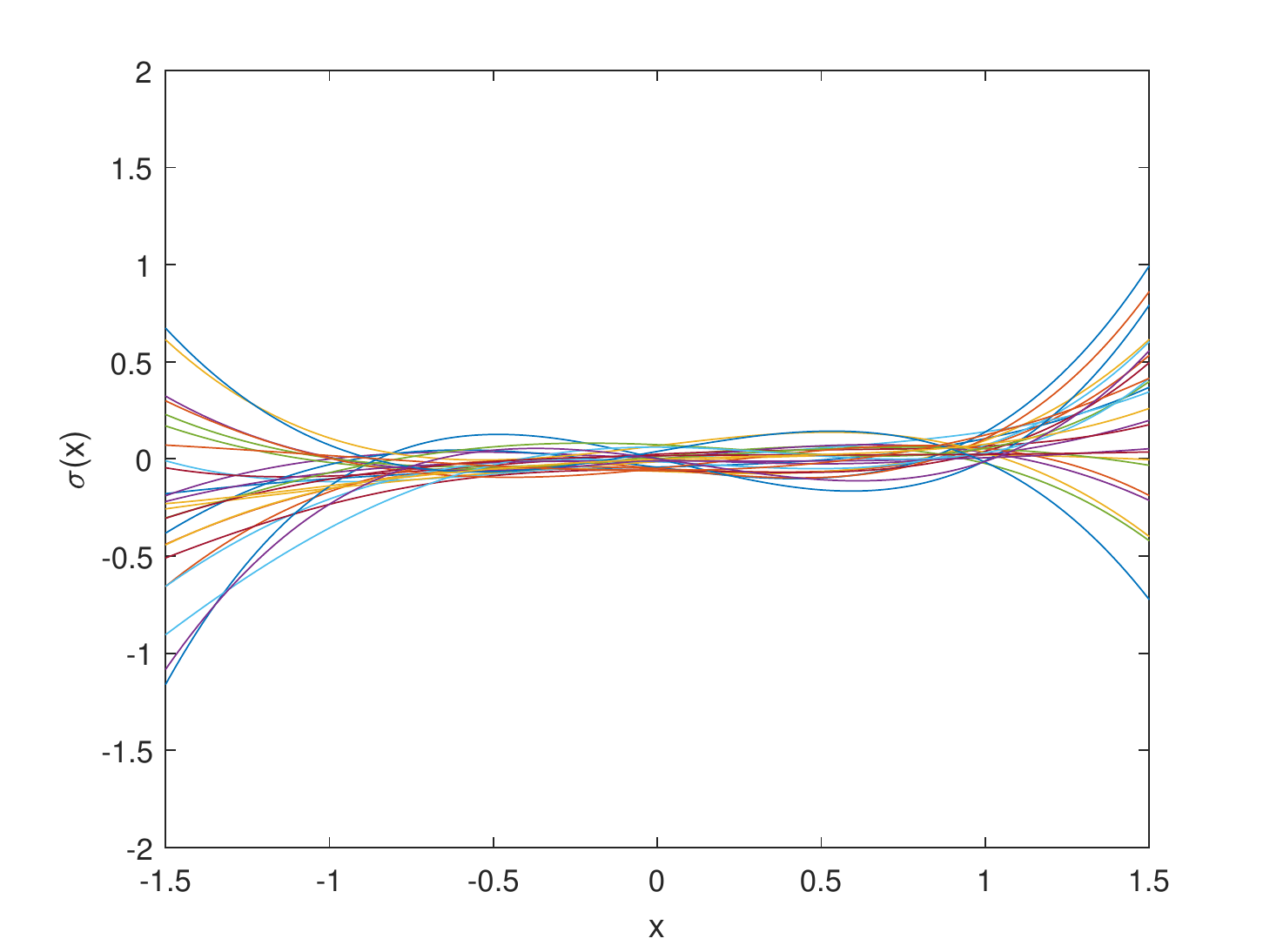}}%
		{\includegraphics[width=0.33\linewidth]{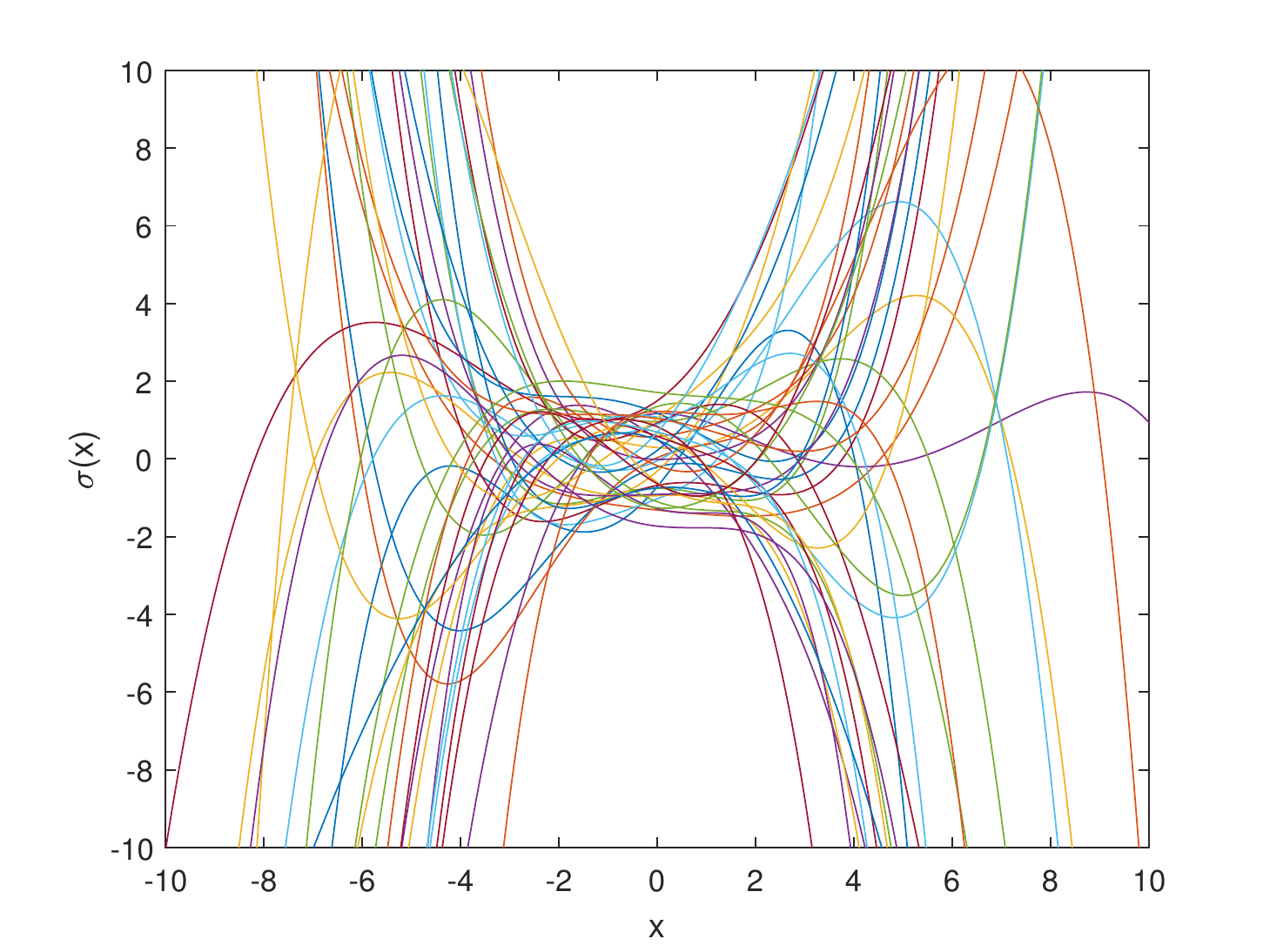}}

		{\footnotesize (a)}
	\end{minipage}%
	
	\begin{minipage}{\linewidth}
		\centering
		{\includegraphics[width=0.33\linewidth]{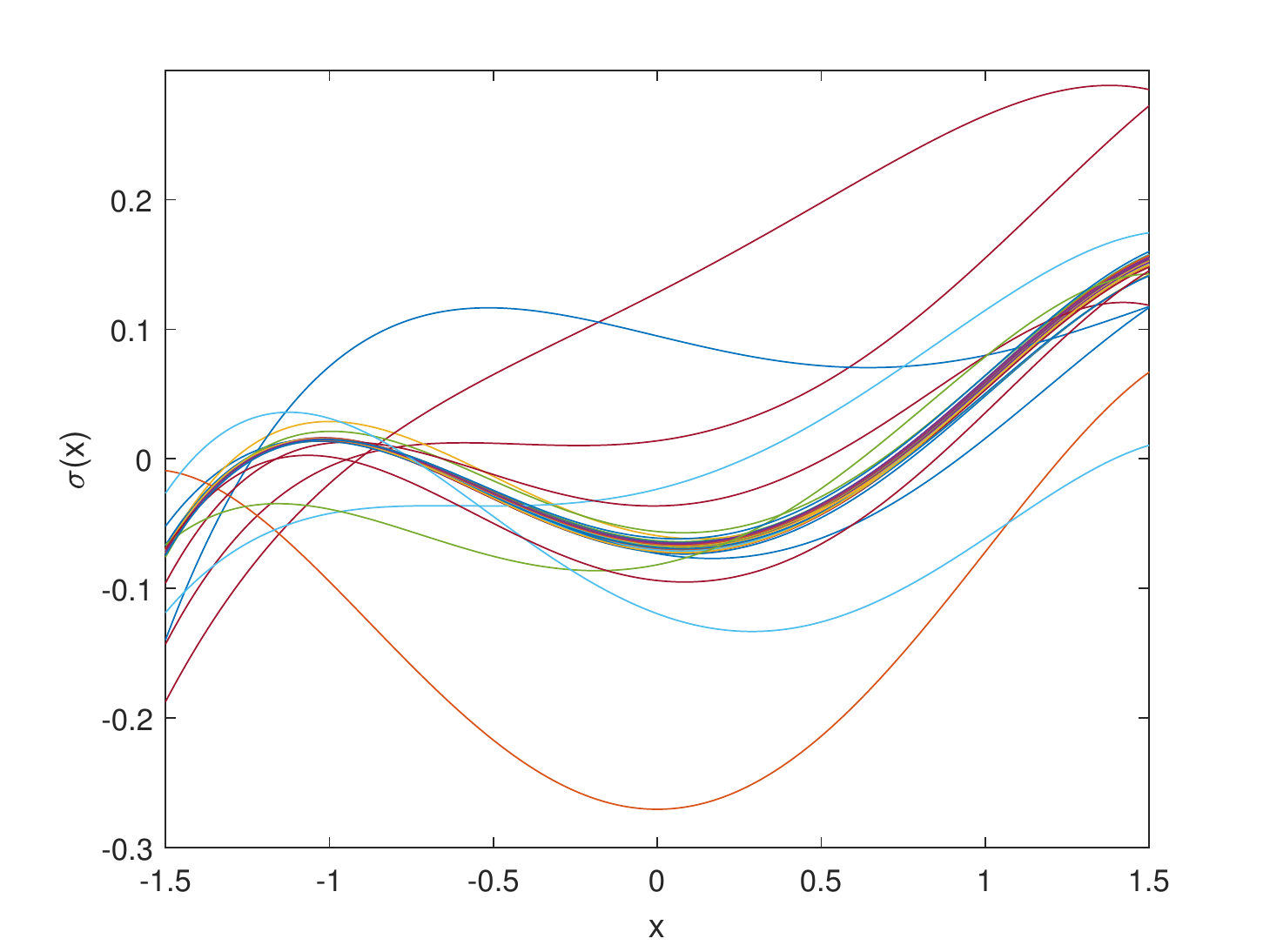}}%
		{\includegraphics[width=0.33\linewidth]{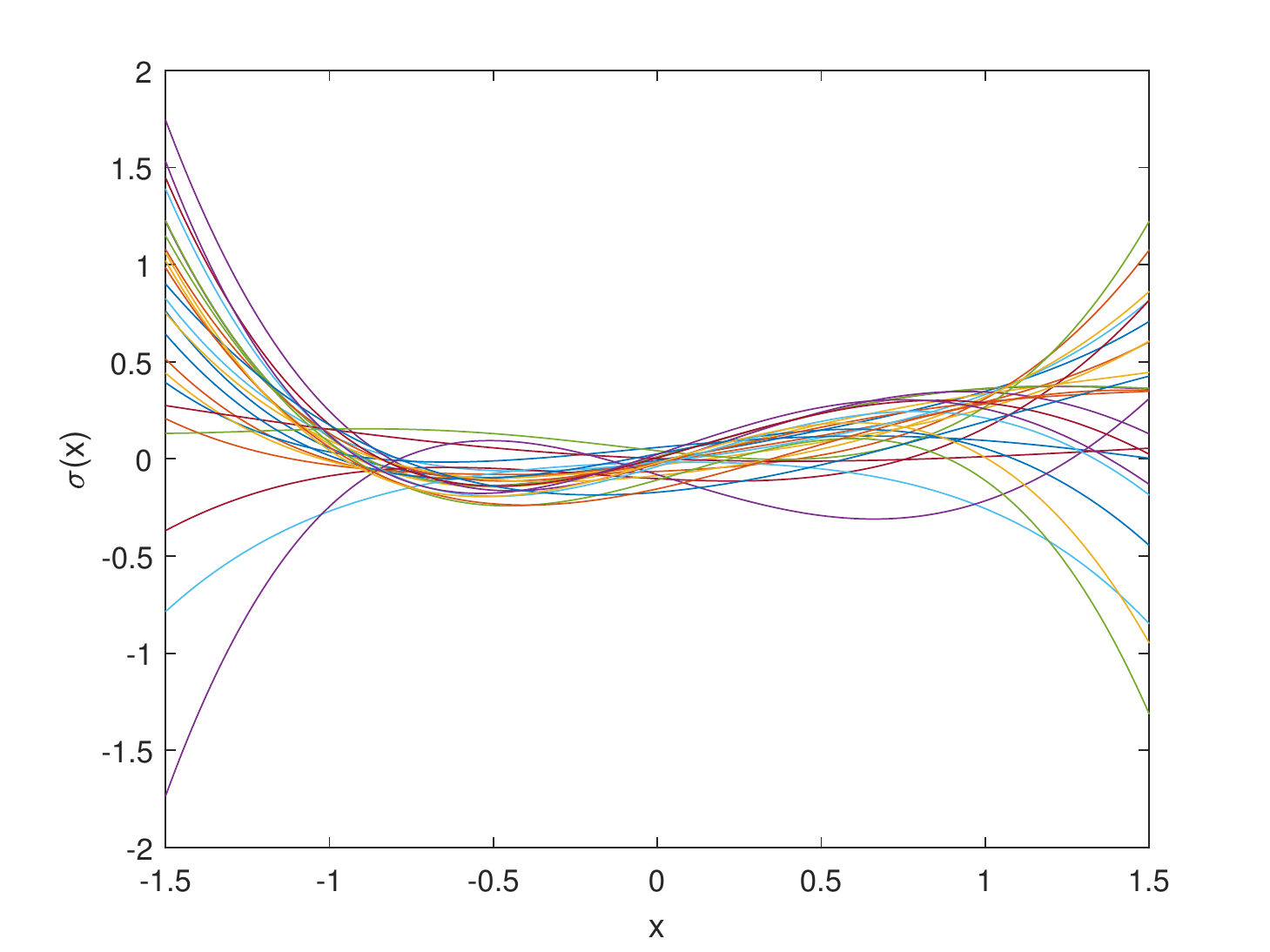}}%
		{\includegraphics[width=0.33\linewidth]{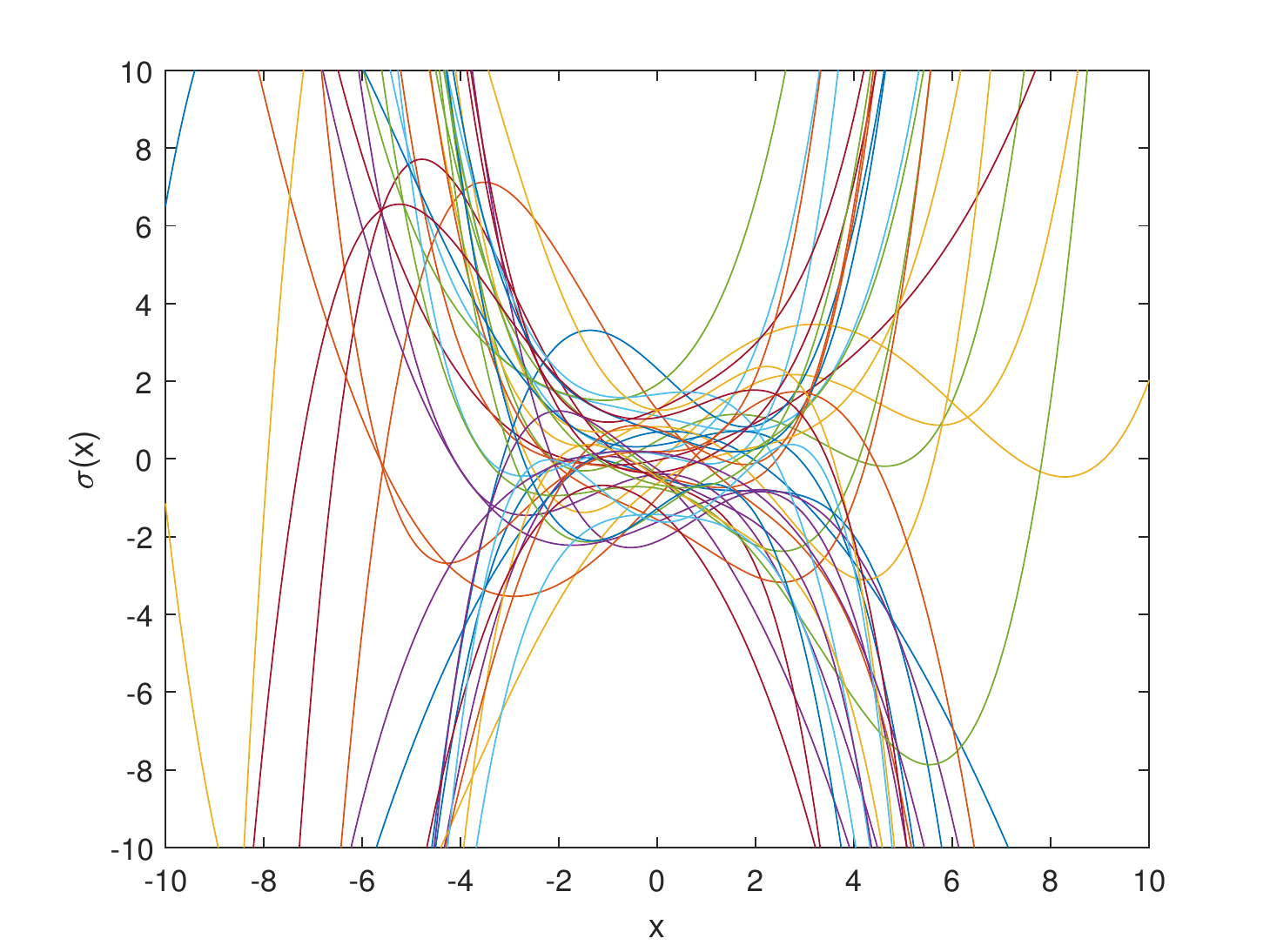}}

		{\footnotesize (b)}
	\end{minipage}
	\caption{Examples of the activation functions in LeNet with activation networks with the CIFAR-10 dataset:
	(a) for dog 1;
	(b) for airplane 1.
	left: the first convolutional layer;
	center: the second convolutional layer;
	right: the first dense layer. 
		}
	\label{fig:lenetact}
\end{figure}

The activation networks provide polynomial coefficients for an activation function of each pixel in a convolution layer and each node in a dense layer. Fig. \ref{fig:lenetact} shows examples of the activation functions used by LeNet-AN for the two input images presented in Fig. \ref{fig:lenetimages}. The activation functions for the pixels in the first and second convolutional layers and those for the nodes in the first dense layer are shown. LeNet-AN uses different shapes of activation functions for its pixels and nodes. Moreover, LeNet-AN adaptively uses different activation functions for the two different inputs. 

\begin{figure}
	\centering
	\begin{minipage}{0.475\linewidth}
	\centering
	\begin{minipage}{0.5\linewidth}
		\centering
		{\includegraphics[trim=64 48 64 25, clip, width=0.5\linewidth]{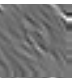}}%
		{\includegraphics[trim=64 48 64 25, clip, width=0.5\linewidth]{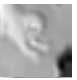}}%

		{\includegraphics[trim=64 48 64 25, clip, width=0.5\linewidth]{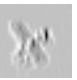}}%
		{\includegraphics[trim=64 48 64 25, clip, width=0.5\linewidth]{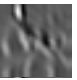}}%
		
		{\footnotesize (a)}
	\end{minipage}%
	\begin{minipage}{0.5\linewidth}
		\centering
		{\includegraphics[trim=64 48 64 25, clip, width=0.5\linewidth]{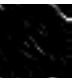}}%
		{\includegraphics[trim=64 48 64 25, clip, width=0.5\linewidth]{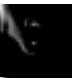}}%

		{\includegraphics[trim=64 48 64 25, clip, width=0.5\linewidth]{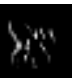}}%
		{\includegraphics[trim=64 48 64 25, clip, width=0.5\linewidth]{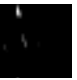}}%

		{\footnotesize (b)}
	\end{minipage}%
	
	\end{minipage}
	\begin{minipage}{0.475\linewidth}
	\centering
	\begin{minipage}{0.5\linewidth}		
		\centering
		{\includegraphics[trim=64 48 64 25, clip, width=0.5\linewidth]{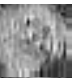}}%
		{\includegraphics[trim=64 48 64 25, clip, width=0.5\linewidth]{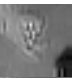}}%

		{\includegraphics[trim=64 48 64 25, clip, width=0.5\linewidth]{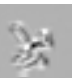}}%
		{\includegraphics[trim=64 48 64 25, clip, width=0.5\linewidth]{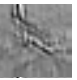}}%

		{\footnotesize (c)}
	\end{minipage}%
	\begin{minipage}{0.5\linewidth}		
		\centering
		{\includegraphics[trim=64 48 64 25, clip, width=0.5\linewidth]{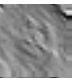}}%
		{\includegraphics[trim=64 48 64 25, clip, width=0.5\linewidth]{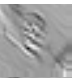}}%

		{\includegraphics[trim=64 48 64 25, clip, width=0.5\linewidth]{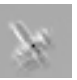}}%
		{\includegraphics[trim=64 48 64 25, clip, width=0.5\linewidth]{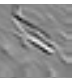}}%

		{\footnotesize (d)}
	\end{minipage}%
	
	\end{minipage}

	\caption{Examples of the activated node outputs for the input images in Fig. \ref{fig:lenetimages}:
	(a) intermediate output of LeNet;
	(b) activated output of LeNet;
	(c) intermediate output of LeNet-AN;
	(d) activated output of LeNet-AN.
	}
	\label{fig:lenetnodeactivation}

\end{figure}

Fig. \ref{fig:lenetnodeactivation} shows examples of activated node outputs of the baseline LeNet in (a) (b) and LeNet-AN in (c) (d), respectively. Intermediate outputs, $u_i^l$, and the corresponding activated outputs, $x_i^l$, of nodes in the first convolutional layer are shown for the input images in Fig. \ref{fig:lenetimages}. The baseline LeNet utilizes directional filters to extract features. The examples show that the single activation function simply passes strong directional features. The examples for LeNet-AN show that the network learns to use filters while relying on selective activation by the adaptive activation functions. The intermediate outputs contain complicated features rather than simple directional features. The activation network adaptively activates only the features suitable for a given task. The activated node outputs show impressions of object being recognized. The use of activation networks enables a network to analyze and adaptively utilize features related to a given task. Node outputs show higher-level features earlier from the shallow layers.

The network performance is reported in Table \ref{tab:lenetcomparison1}. LeNet-AN provides higher accuracy than the baseline LeNet. However, the use of activation networks increases the number of parameters in a network. As a result, we use a more complex network than the baseline network for a given task. For fair comparison, deeper versions of LeNet with more layers and wider versions of LeNet with more nodes in layers are prepared. The networks are trained using the same training set. The test accuracy and number of parameters of the baseline, deeper, and wider LeNets are reported in Table \ref{tab:lenetcomparison1}. The number of parameters of LeNet-AN is only 121.5\% of that of the baseline LeNet. However, LeNet-AN provides higher accuracy than the more complex deeper versions of LeNets. The wider versions 1 and 2 of LeNet provide higher accuracy, but their complexity is 392.5\% and 877.4\% of the baseline LeNet, respectively. We prepare a deeper version of LeNet with activation networks. The number of parameters of the deeper LeNet-AN is only 143.6\% of that of the baseline LeNet. However, it provides higher accuracy than the much more complex wider versions of LeNets.

\begin{table}[h!]										
\centering										
\caption{Performance vs. Parameters for LeNet using CIFAR-10}														
\label{tab:lenetcomparison1}
\begin{tabular}{ll|rr|c}	
	\hline
	\multicolumn{2}{c|}{network} & \multicolumn{2}{c|}{\# of parameters}  & test accuracy\\	\hline
	LeNet & baseline & 37406 & (100.0\%) & 0.6900 \\ \cline{2-5}
	& deeper 1 & 49994 & (133.7\%) & 0.7093 \\ \cline{2-5}
	& deeper 2 & 64894 & (173.5\%) & 0.7119 \\ \cline{2-5}
	& wider 1 & 146802 & (392.5\%) & 0.7407 \\ \cline{2-5}
	& wider 2 & 328198 & (877.4\%) & 0.7540 \\ \hline
	LeNet-AN & baseline & 45456 & (121.5\%) & 0.7331 \\ \cline{2-5}							
	& deeper 1 & 53708 & (143.6\%) & 0.7791\\ \hline							
\end{tabular}
\end{table}

\begin{figure}[h]
	\centering
	\centering
	\begin{minipage}{0.25\linewidth}
		\centering
		{\includegraphics[width=\linewidth]{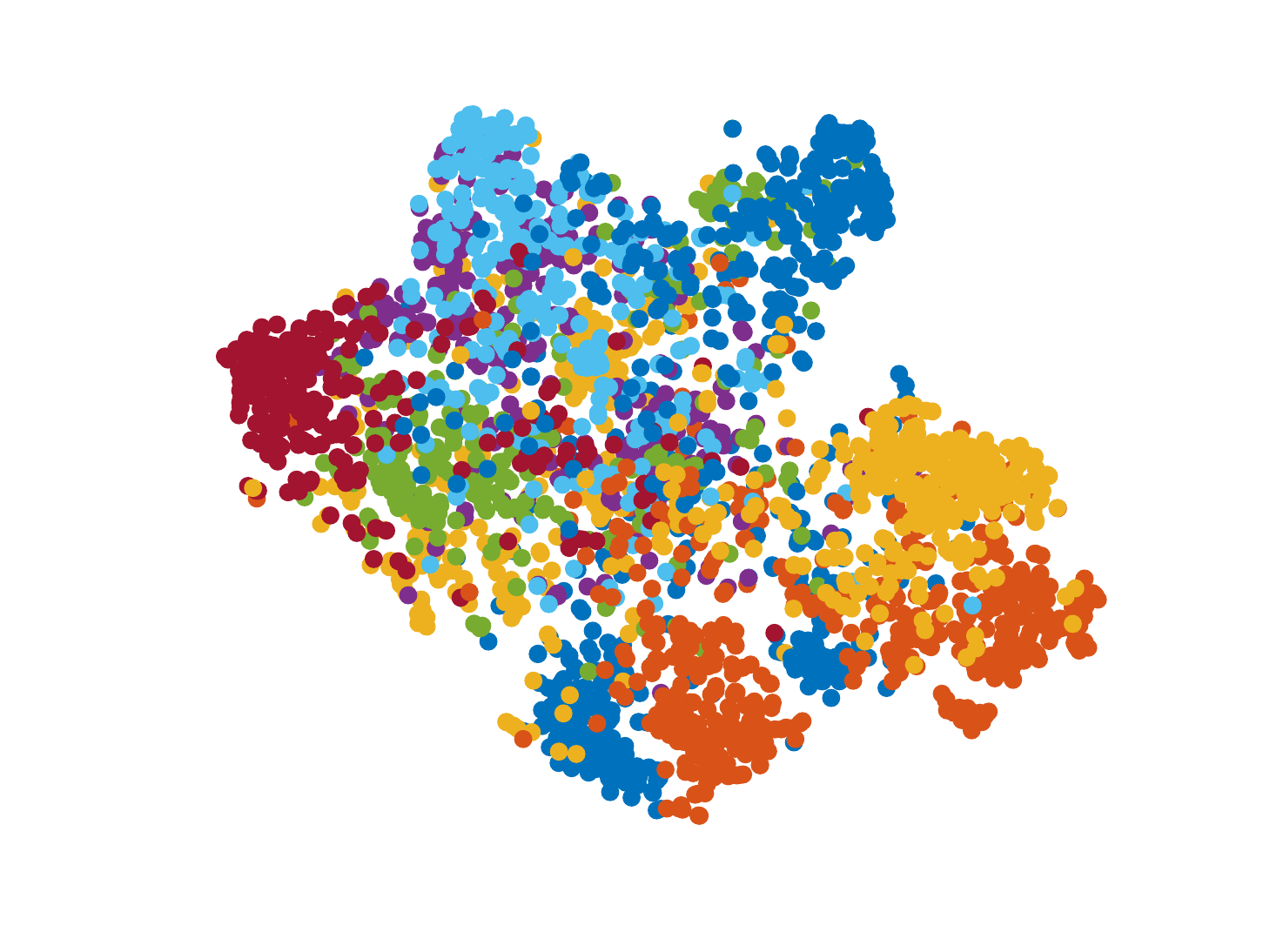}}%

		{\footnotesize (a)}
	\end{minipage}%
	\begin{minipage}{0.25\linewidth}
		\centering
		{\includegraphics[width=\linewidth]{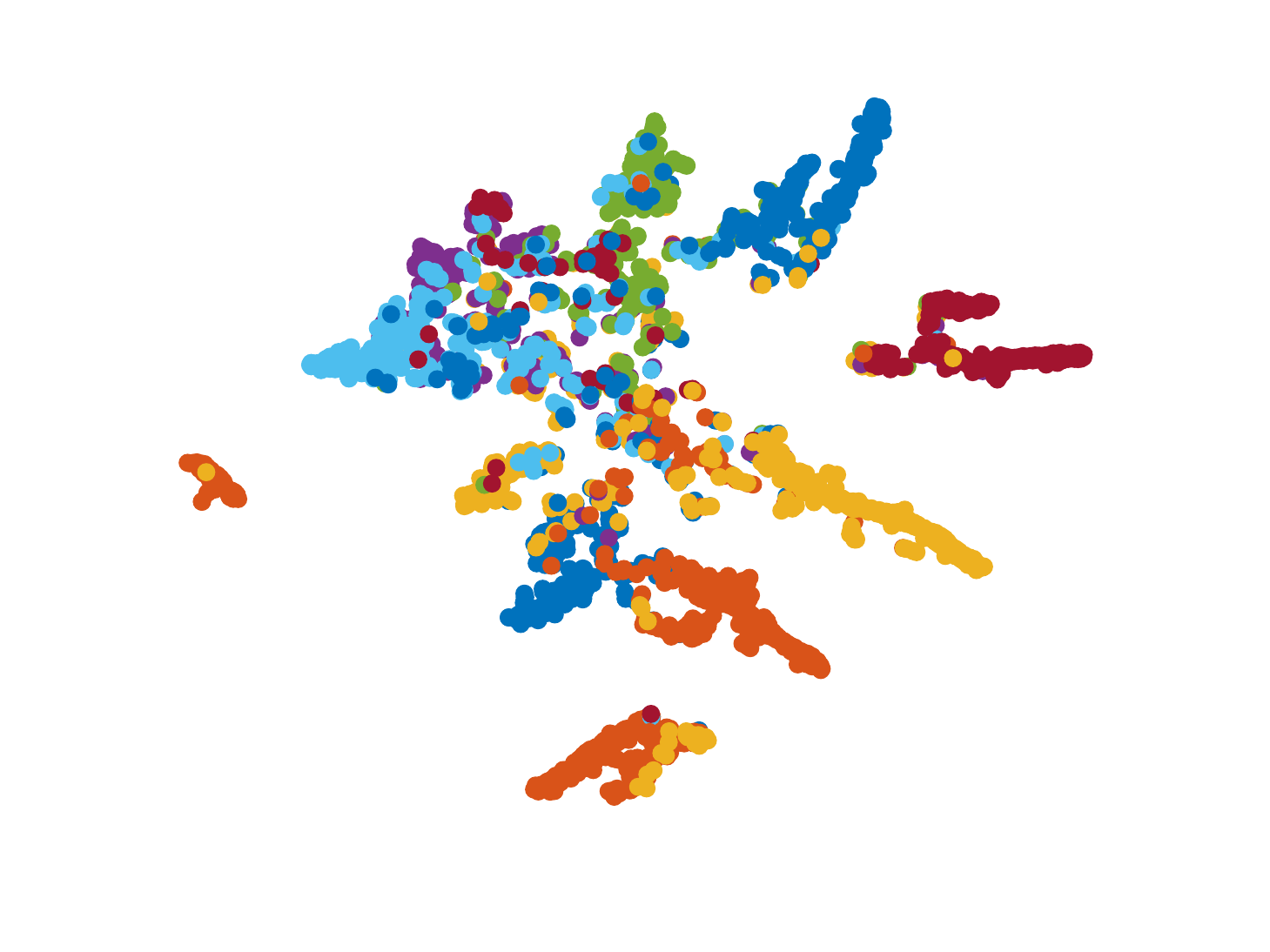}}%
		
		{\footnotesize (b)}
	\end{minipage}%
	\begin{minipage}{0.25\linewidth}
		\centering
		{\includegraphics[width=\linewidth]{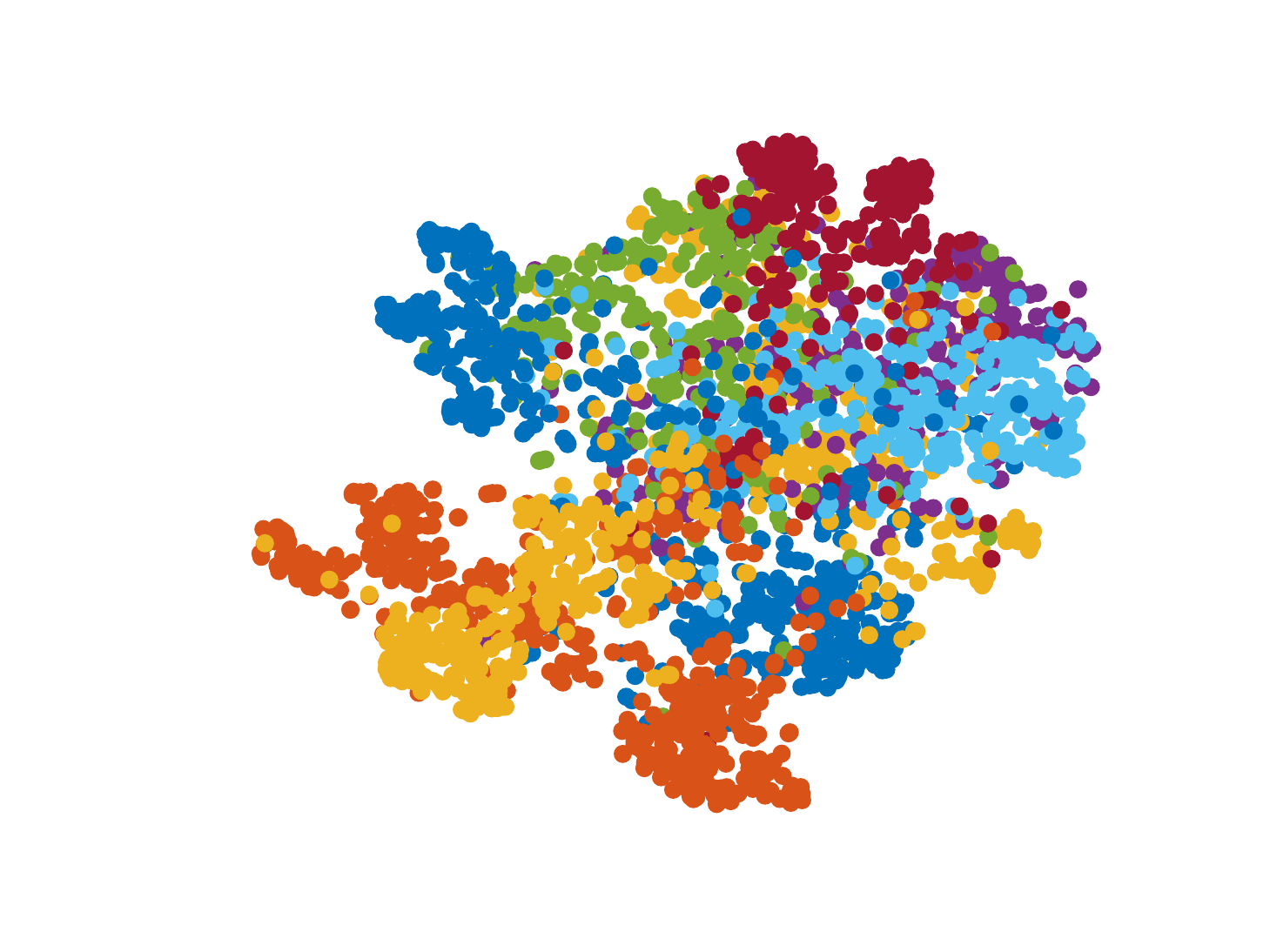}}%

		{\footnotesize (c)}
	\end{minipage}%
	\begin{minipage}{0.25\linewidth}
		\centering
		{\includegraphics[width=\linewidth]{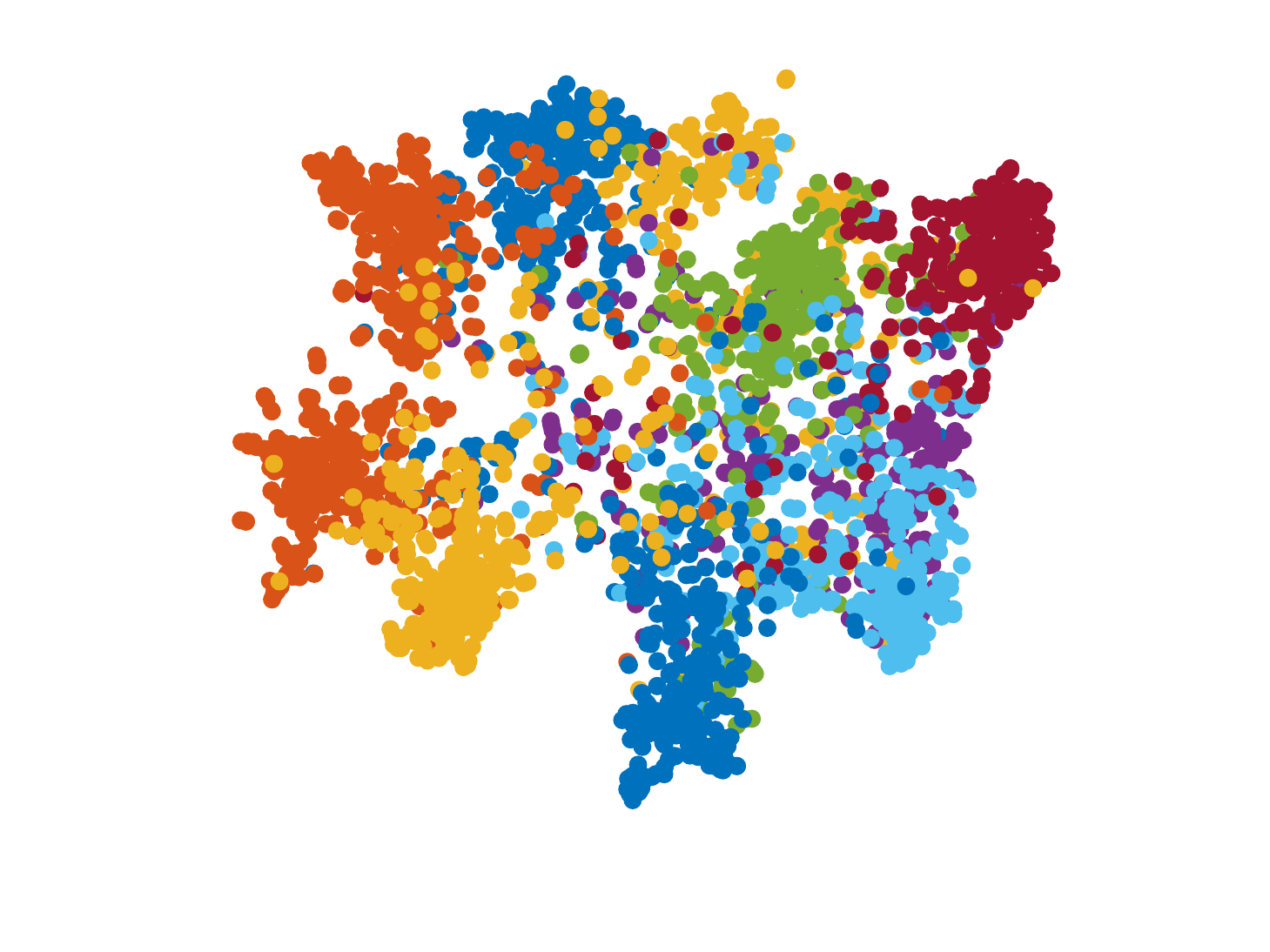}}%
		
		{\footnotesize (d)}
	\end{minipage}%
	\caption{t-SNE of features used by networks, LeNet using CIFAR-10,
	features in the last dense layer:
	(a) baseline LeNet;
	(b) deeper LeNet;
	(c) wider LeNet;
	(d) LeNet-AN.
	}
	\label{fig:tsne}
\end{figure}

Fig. \ref{fig:tsne} shows the t-SNE analysis of the features in the last dense layer of the baseline, deeper, wider LeNets and LeNet-AN. The features of objects are better separated by LeNet-AN than by the baseline and deeper LeNets and comparable to the much complex wider LeNet, which confirms the improved performance of LeNet-AN.

\begin{figure}[h]
	\centering
	\begin{minipage}{0.4\linewidth}
		\centering
		{\includegraphics[width=\linewidth]{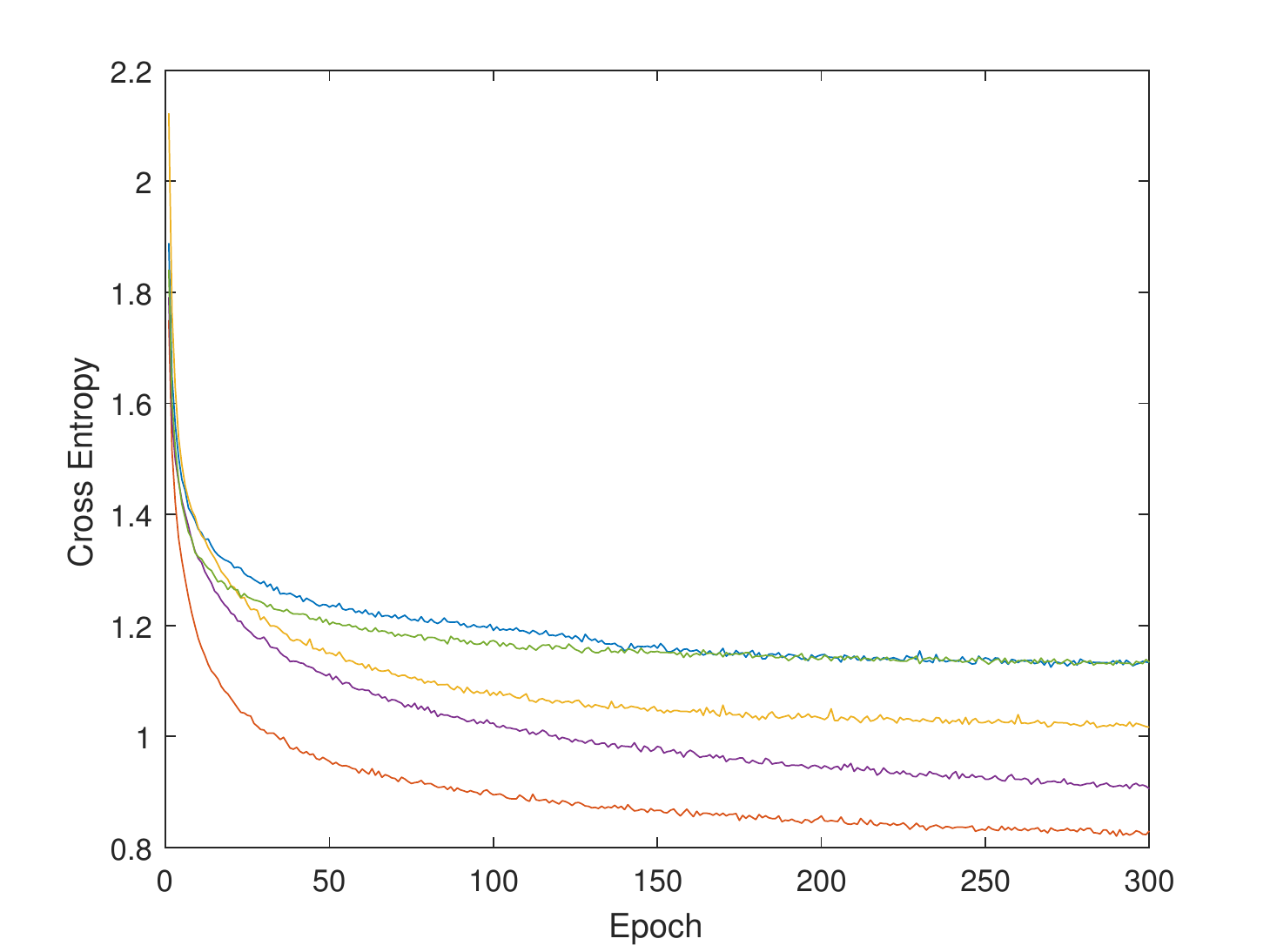}}%

		{\footnotesize (a)}
	\end{minipage}%
	\begin{minipage}{0.4\linewidth}
		\centering
		{\includegraphics[width=\linewidth]{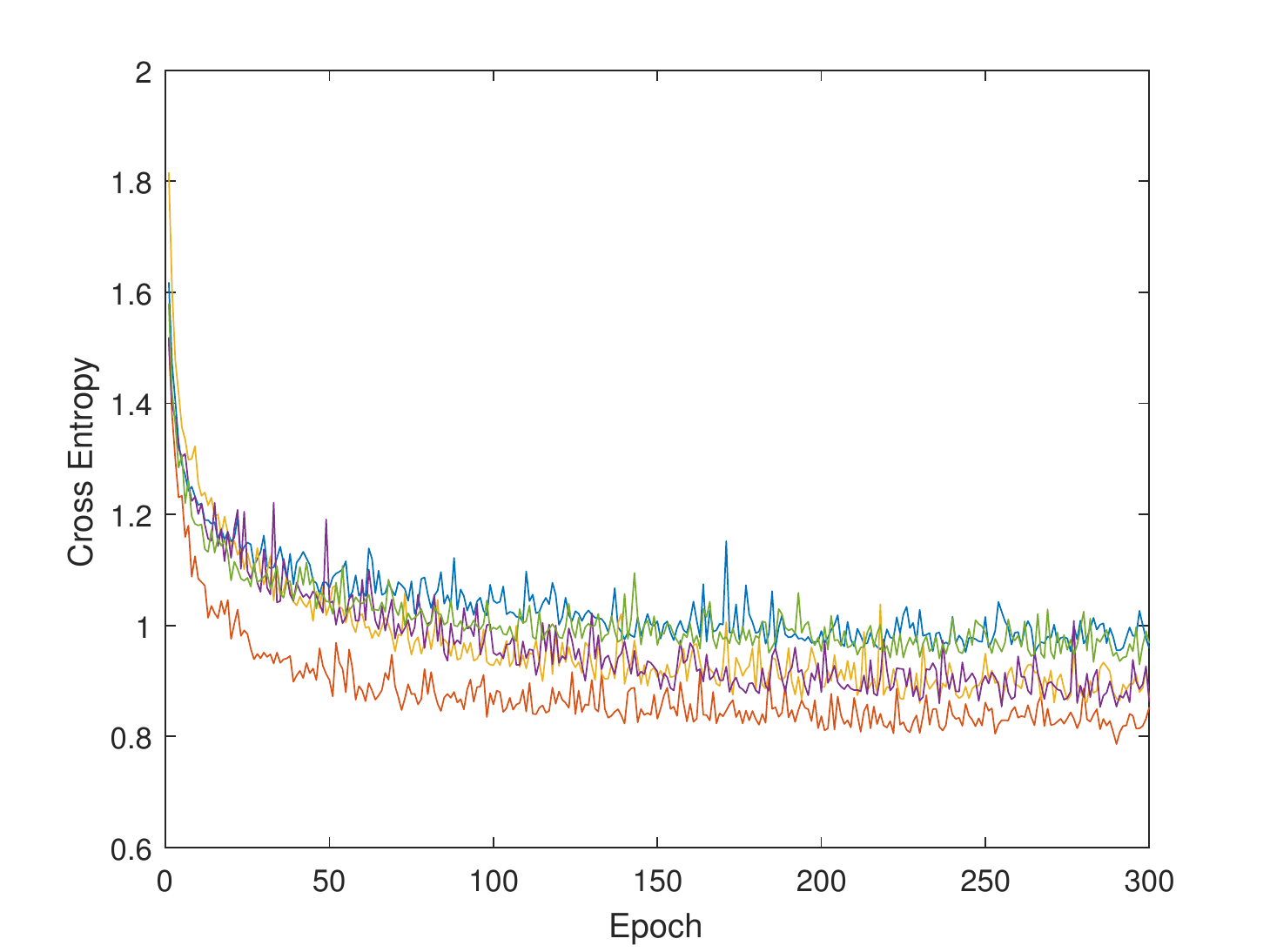}}%
		
		{\footnotesize (b)}
	\end{minipage}%
	
	\caption{Training of networks with various activation methods: LeNet with CIFAR-10.
		(a) training loss;
		(b) validation loss.
		blue: baseline LeNet; 
		yellow: LeNets with polynomial function;
		green: LeNets with inhibition model;
		purple: LeNets with attention model;
		red: LeNet-AN.
		}
	\label{fig:lenetcomparison}
\end{figure}

We compared the network performance to other activation methods. For comparison, the polynomial activation with polynomial coefficients learned using the training set \cite{piazza1992artificial, chung2016deep}, the inhibition model that modifies the activation with the sum of neighboring node outputs \cite{fernandes2013lateral}, and the attention models that modulate the activation with the attention learned with a simple network \cite{xuk2015show} are prepared for LeNet. Fig. \ref{fig:lenetcomparison} shows the training and validation losses for the networks with different activation methods. The proposed network was trained much faster with a lower training loss than the other networks. Table \ref{tab:lenetcomparison2} shows the performance and number of parameters of the networks. For object recognition, all networks with variant activation methods provided higher accuracy than the baseline LeNet. The proposed method outperformed all methods in providing the highest accuracy at a modest increase in complexity.

\begin{table}[h!]										
\centering										
\caption{Comparisons of the Network Performance for LeNet with CIFAR-10}	
\label{tab:lenetcomparison2}																		
\begin{tabular}{ll|cc|c}	
	\hline
	\multicolumn{2}{c|}{network} & \multicolumn{2}{c|}{\# of parameters} & test accuracy \\	\hline
	LeNet & baseline & 37406 & (100.0\%)& 67.13\%\\ \cline{2-5}
	& polynomial & 37936 & (101.4\%) & 70.12\%\\ \cline{2-5}
	& inhibition & 37406 & (100.0\%)& 67.68\%\\ \cline{2-5}
	& attention & 44860 & (119.9\%)& 70.69\%\\ \hline
	LeNet-AN & & 45456 & (121.5\%)& 72.56\%\\ \hline							
\end{tabular}
\end{table}

\subsection{U-net for Denoising}
\label{sec:unet}

\begin{figure}[h]
	\centering
	\begin{minipage}{0.4\linewidth}
		\centering
		{\includegraphics[width=\linewidth]{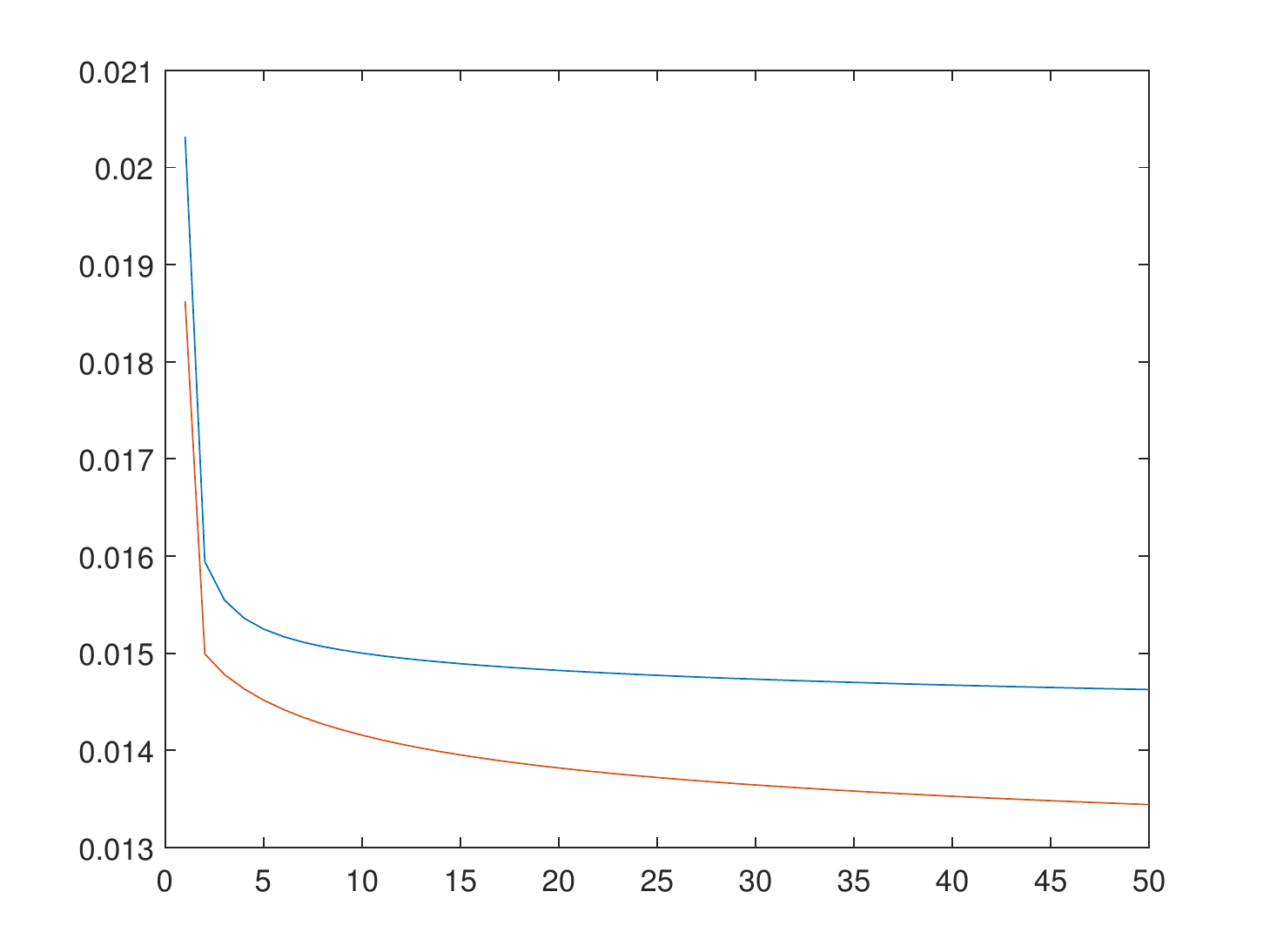}}%
		
		{\footnotesize (a)}
	\end{minipage}%
	\begin{minipage}{0.4\linewidth}
		\centering
		{\includegraphics[width=\linewidth]{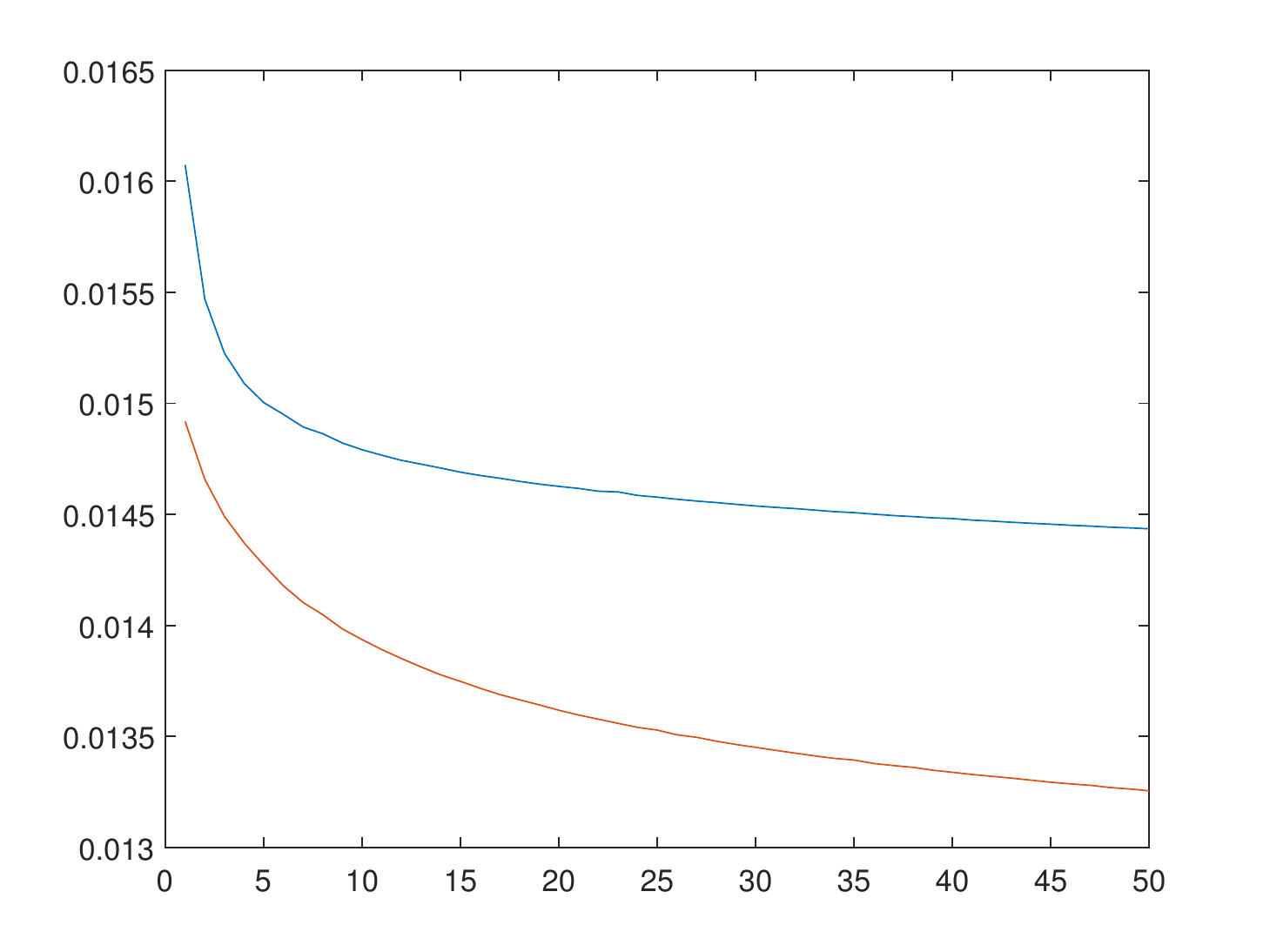}}%
				
		{\footnotesize (b)}
	\end{minipage}%
	
	\caption{Training of U-net with MNIST:
		(a) training loss;
		(b) validation loss.
		blue: baseline U-net;
		red: U-net-AN.
		}
	\label{fig:unettraining}
\end{figure}

We prepared U-net \cite{ronneberger2015u} by adding the proposed activation network for each convolutional layer for denoising. A baseline U-net with the ReLU activation function for all nodes was also prepared for comparison. The U-net with activation network is denoted by U-net-AN. The variance of noise added to the ground truth images to prepare noisy input images is 0.05. The networks were trained using the MNIST dataset \cite{lecun1998gradient}. Fig. \ref{fig:unettraining} shows the training and validation losses for the baseline U-net and U-net-AN. U-net-AN was trained much faster with lower training and validation losses. 

Table \ref{tab:unetcomparison} shows the mean squared error (MSE) between the denoised images and their ground truth images. U-net-AN provides an improvement in MSE.

\begin{table}[h!]										
\centering										
\caption{Performance vs. Parameters for U-net with MNIST}	
\label{tab:unetcomparison}																		
\begin{tabular}{l|cc|c}	
	\hline
	network & \multicolumn{2}{c|}{\# of parameters} & MSE \\	\hline
	baseline & 231041& (100.0\%) & 0.0144\\ \hline
	proposed & 365761 & (158.3\%) & 0.0133\\ \hline							
\end{tabular}
\end{table}	

\begin{figure}[h]
	\centering
	\begin{minipage}{0.25\linewidth}
		\centering
		{\includegraphics[trim=520 345 530 180, clip, width=\linewidth]{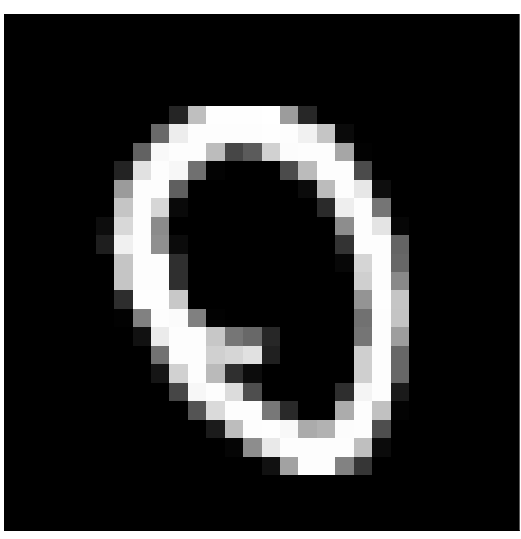}}%
		
		{\footnotesize (a)}
	\end{minipage}%
	\begin{minipage}{0.25\linewidth}
		\centering		
		{\includegraphics[trim=520 345 530 180, clip, width=\linewidth]{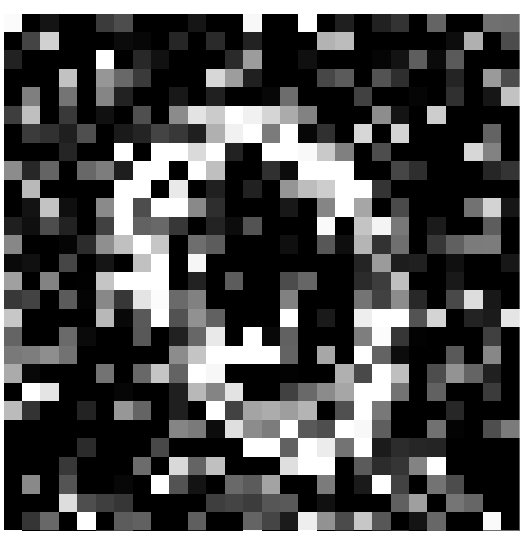}}

		{\footnotesize (b)}
	\end{minipage}%
	\begin{minipage}{0.25\linewidth}
		\centering
		{\includegraphics[trim=520 345 530 180, clip, width=\linewidth]{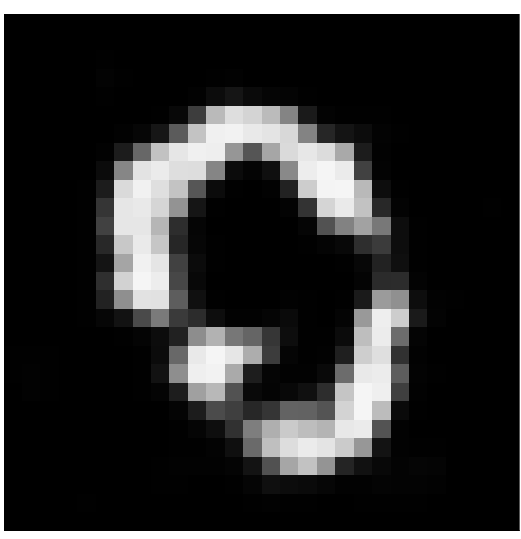}}%
		
		{\footnotesize (c)}
	\end{minipage}%
	\begin{minipage}{0.25\linewidth}
		\centering
		{\includegraphics[trim=520 345 530 180, clip, width=\linewidth]{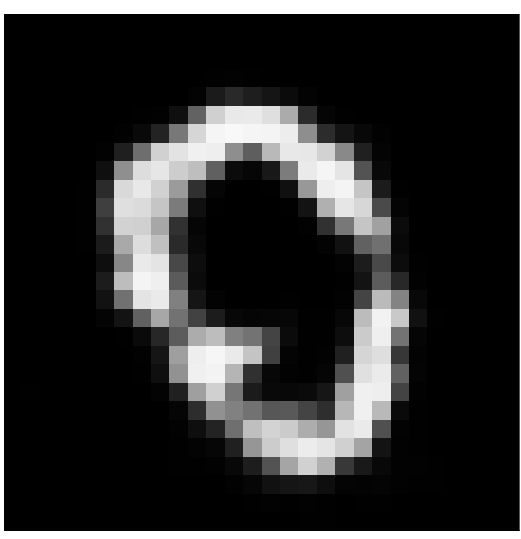}}%
		
		{\footnotesize (d)}
	\end{minipage}
	\caption{Examples of denoising by U-net:
	(a) ground truth image;
	(b) noisy input image;
	followed by images denoised by
	(c) baseline U-net;
	(d) U-net-AN.
	}
	\label{fig:unetresults}
\end{figure}

Fig. \ref{fig:unetresults} shows examples of denoised images. The original and noisy images are shown in Fig. \ref{fig:unetresults} (a) and (b), respectively. The letter zero image denoised by U-net-AN is much closer to the ground truth than the image denoised by the baseline U-net. The letter does not have as many disconnections in the stroke in Fig. \ref{fig:unetresults}  (d) as that in Fig. \ref{fig:unetresults} (c).

\begin{figure}[h]
	\centering
	\begin{minipage}{\linewidth}
		\centering
		{\includegraphics[trim=510 340 515 160, clip, width=0.2\linewidth]{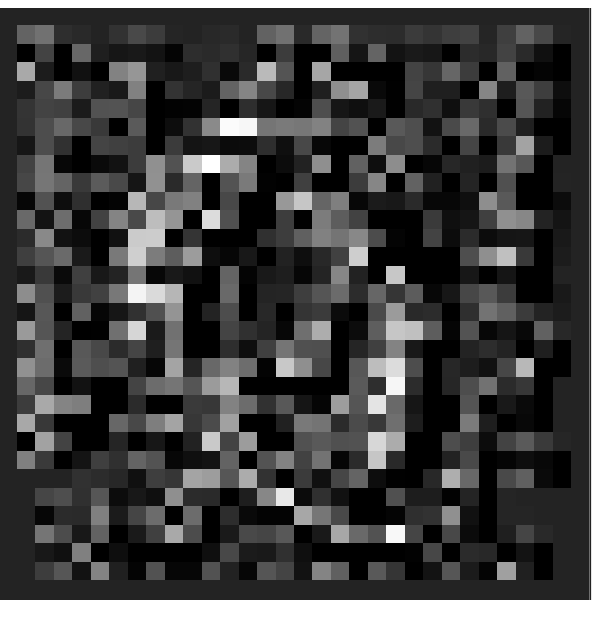}}%
		{\includegraphics[trim=510 340 515 160, clip, width=0.2\linewidth]{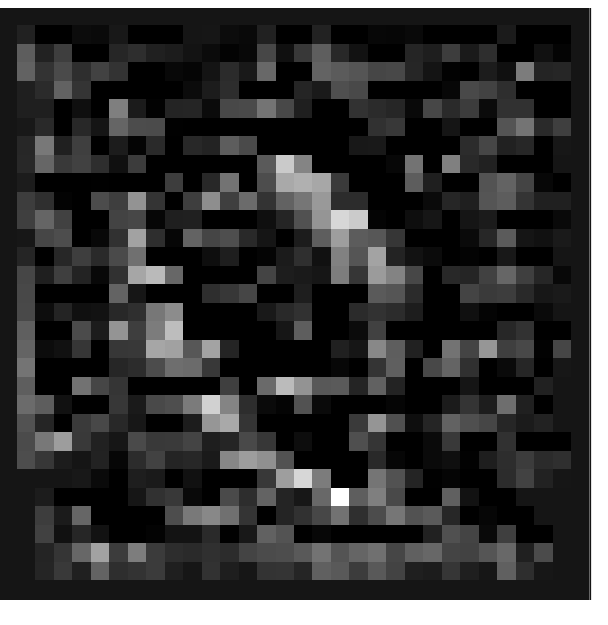}}%
		{\includegraphics[trim=510 340 515 160, clip, width=0.2\linewidth]{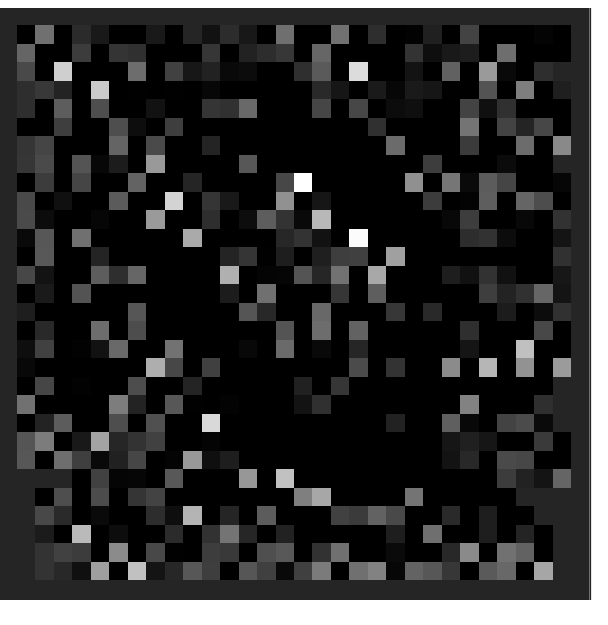}}%
		{\includegraphics[trim=510 340 515 160, clip, width=0.2\linewidth]{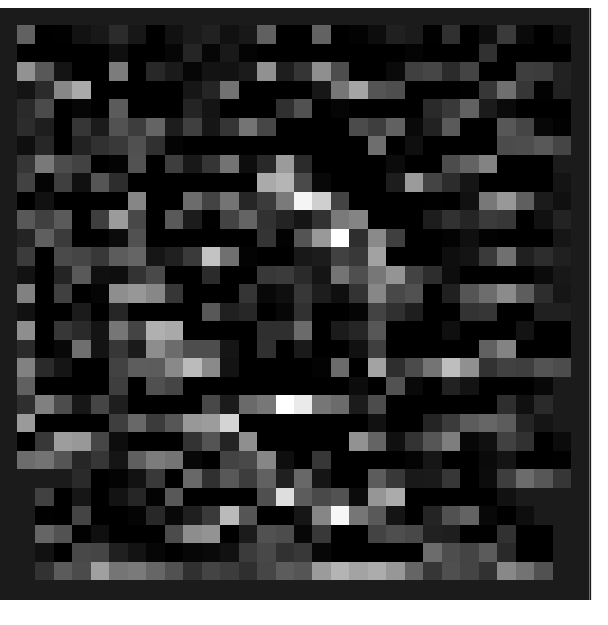}}%
		{\includegraphics[trim=510 340 515 160, clip, width=0.2\linewidth]{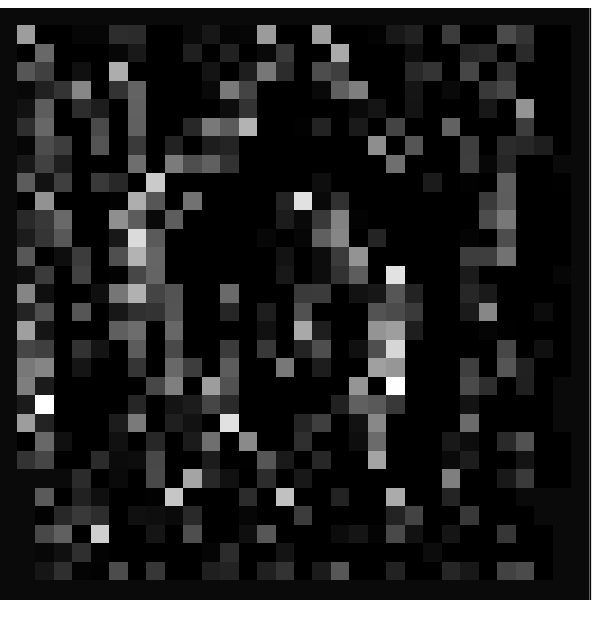}}%
		
		{\includegraphics[trim=63.8 48.2 64.7 35.2, clip, width=0.1\linewidth]{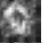}}%
		{\includegraphics[trim=63.8 48.2 64.7 35.2, clip, width=0.1\linewidth]{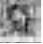}}%
		{\includegraphics[trim=63.8 48.2 64.7 35.2, clip, width=0.1\linewidth]{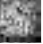}}%
		{\includegraphics[trim=63.8 48.2 64.7 35.2, clip, width=0.1\linewidth]{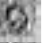}}%
		{\includegraphics[trim=63.8 48.2 64.7 35.2, clip, width=0.1\linewidth]{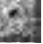}}%
		
		{\footnotesize (a)}
	\end{minipage}%
	
	\begin{minipage}{\linewidth}
		\centering
		{\includegraphics[trim=510 340 515 160, clip, width=0.2\linewidth]{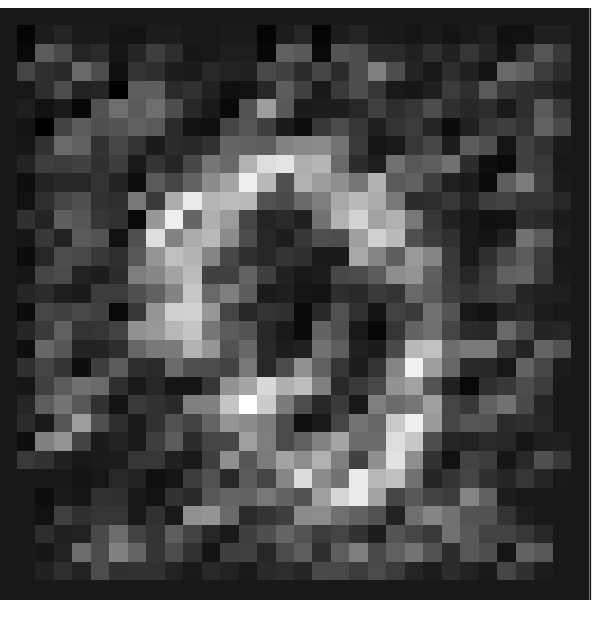}}%
		{\includegraphics[trim=510 340 515 160, clip, width=0.2\linewidth]{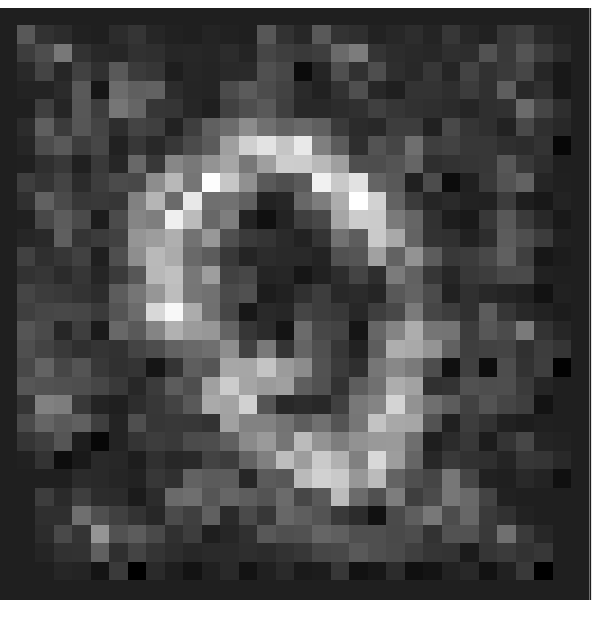}}%
		{\includegraphics[trim=510 340 515 160, clip, width=0.2\linewidth]{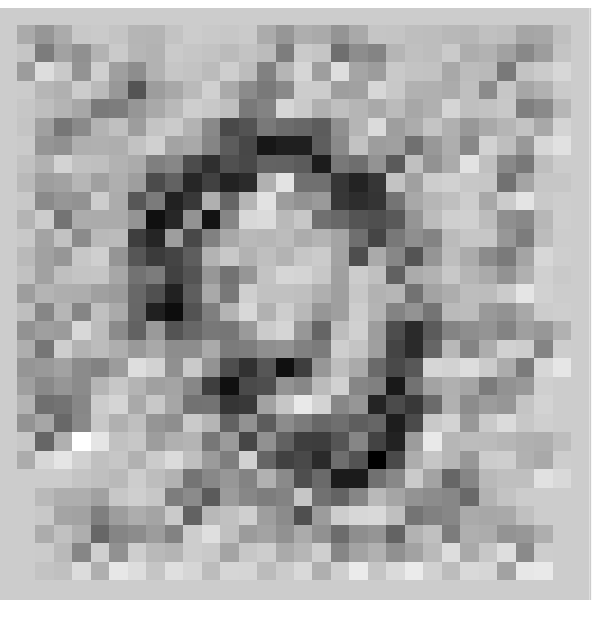}}%
		{\includegraphics[trim=510 340 515 160, clip, width=0.2\linewidth]{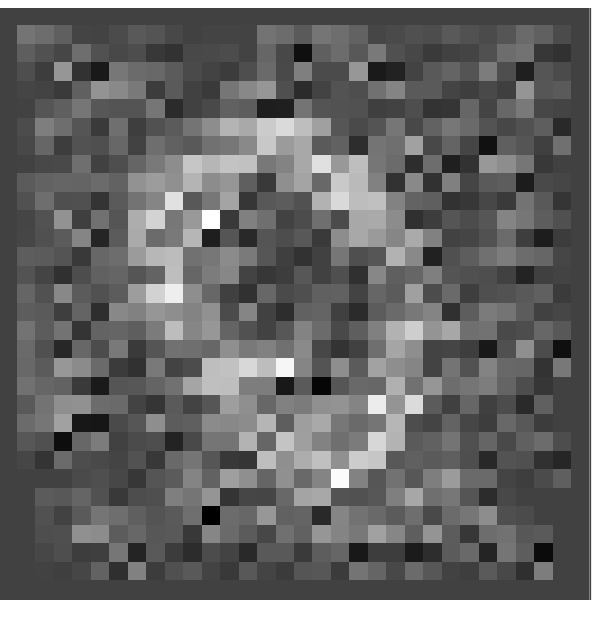}}%
		{\includegraphics[trim=510 340 515 160, clip, width=0.2\linewidth]{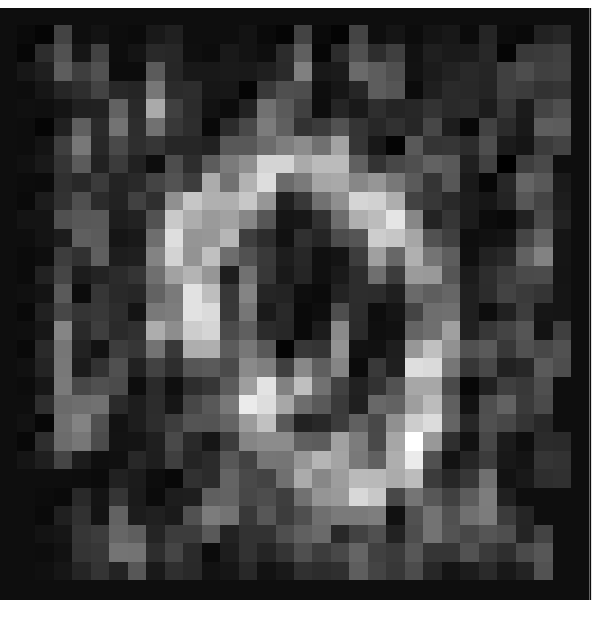}}%
		
		{\includegraphics[trim=63.8 48.2 64.7 35.2, clip, width=0.1\linewidth]{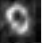}}%
		{\includegraphics[trim=63.8 48.2 64.7 35.2, clip, width=0.1\linewidth]{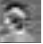}}%
		{\includegraphics[trim=63.8 48.2 64.7 35.2, clip, width=0.1\linewidth]{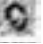}}%
		{\includegraphics[trim=63.8 48.2 64.7 35.2, clip, width=0.1\linewidth]{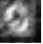}}%
		{\includegraphics[trim=63.8 48.2 64.7 35.2, clip, width=0.1\linewidth]{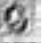}}%
		
		{\footnotesize (b)}
	\end{minipage}%
	\caption{Randomly selected node outputs of U-net:
	(a) baseline U-net;
	(b) U-net-AN.
	top: the first convolutional layer;
	bottom: the second convolutional layer.
	}
	\label{fig:unetfeature}
\end{figure}

Examples of the features used by the U-nets are shown in Fig. \ref{fig:unetfeature}. The node outputs of the first and second convolutional layers are shown. The node outputs of the baseline U-net in Fig. \ref{fig:unetfeature} (a) show that the features in the first layer are mostly directional features. The features are affected by the noise in the input image. There are strong node outputs even where there are no directional features in the ground truth image. High-level features appear in the second layer. The high-level features are affected by the noise resides in the low-level features. The node outputs of U-net-AN are shown in Fig. \ref{fig:unetfeature} (b). The proposed network uses high-level features from the first layers. The impressions of the letter zero are clearly observed in the node outputs. The features in the second layer show a much more effective removal of noise.

\begin{figure}[h]
	\centering
	
	\begin{minipage}{0.4\linewidth}
		\centering
		
		{\includegraphics[width=\linewidth]{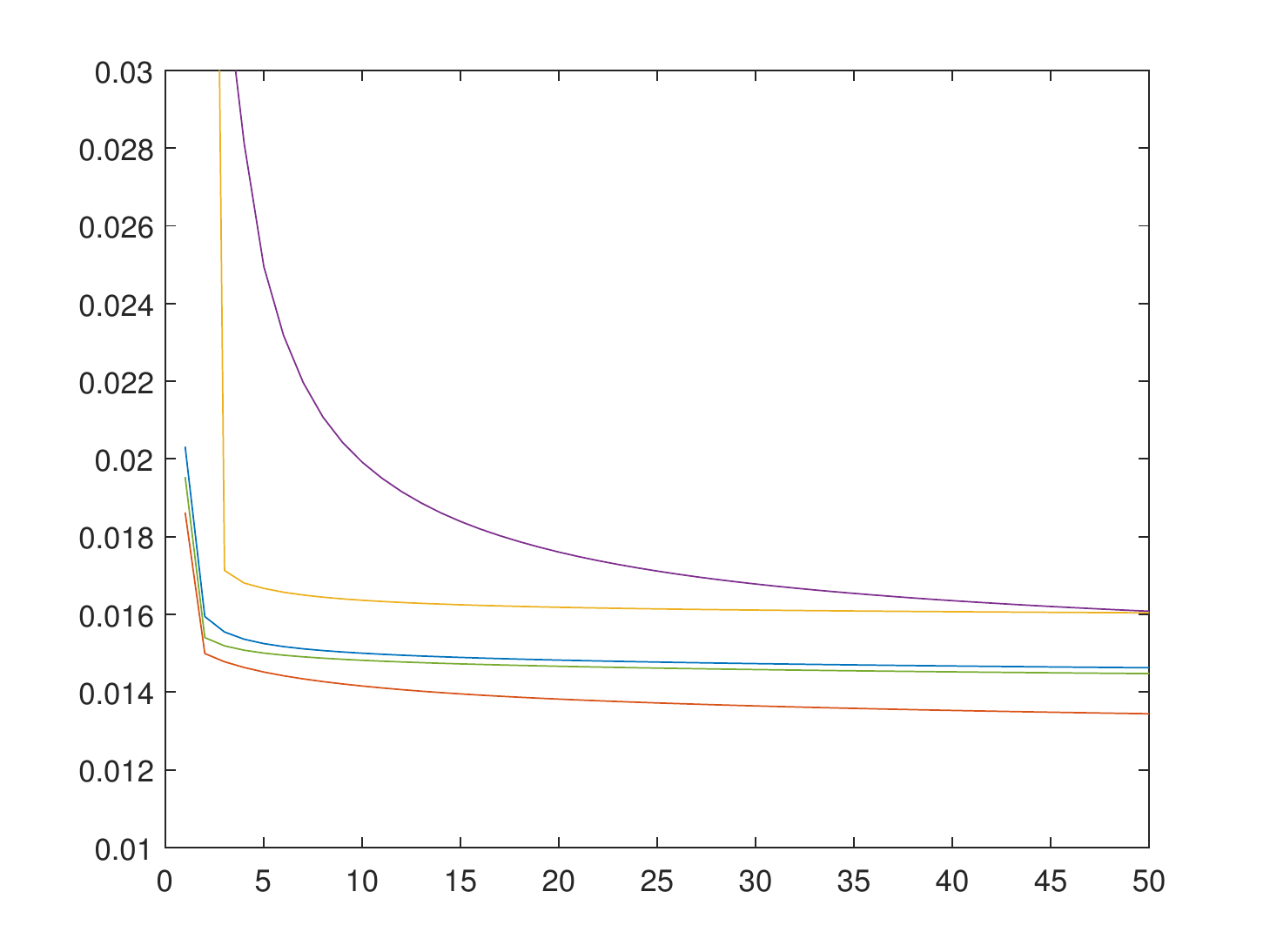}}%

		{\footnotesize (a)}
	\end{minipage}%
	\begin{minipage}{0.4\linewidth}
		\centering
		
		{\includegraphics[width=\linewidth]{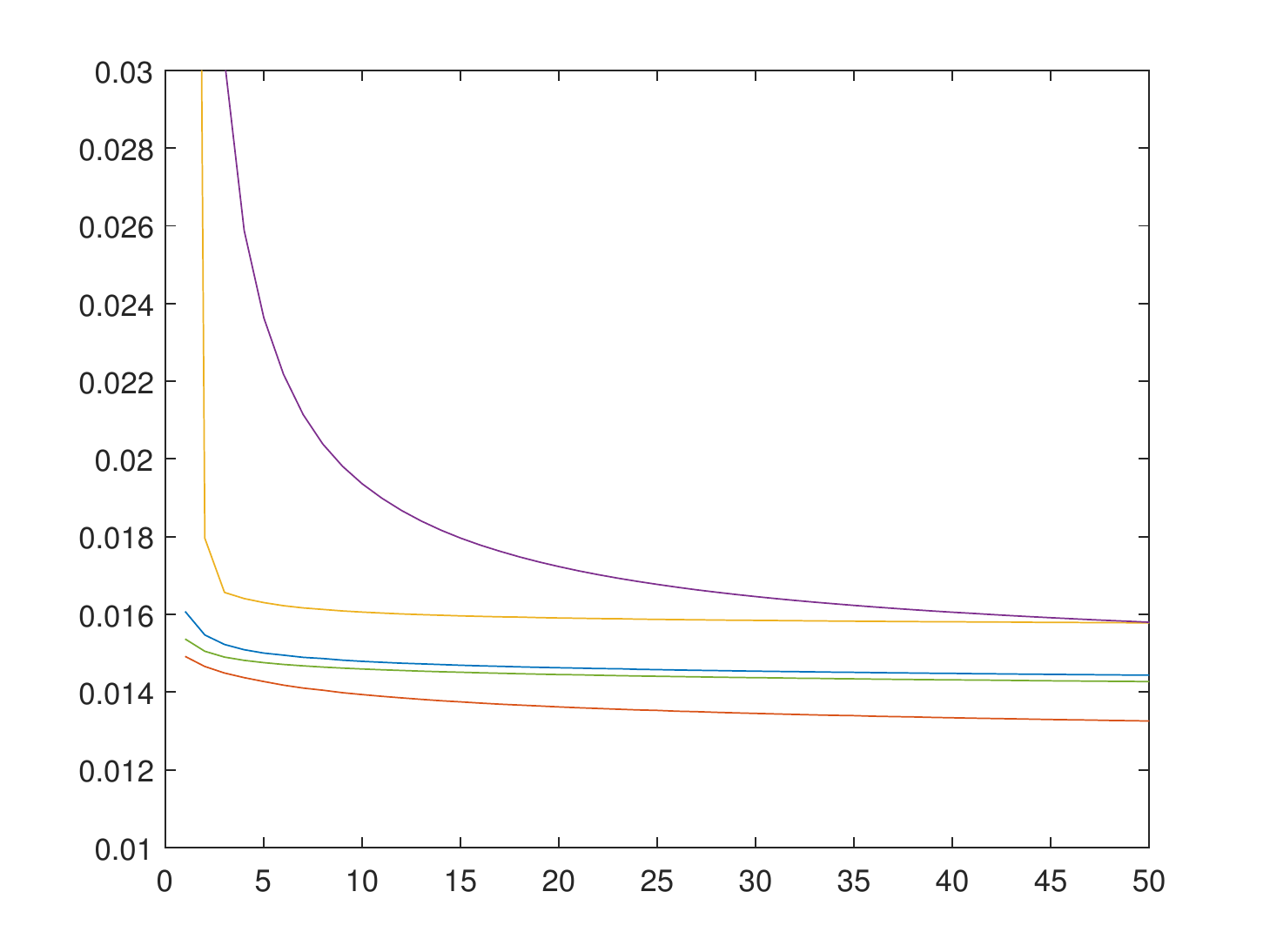}}%

		{\footnotesize (b)}
	\end{minipage}%
	\caption{Training of networks with various activation methods, U-net with MNIST.
		(a) training loss and
		(b) validation loss.
		blue: baseline LeNet, 
		U-net with 
		yellow: polynomial function, 
		green: inhibition model,
		purple: attention model,
		red: U-net-AN.
		}
	\label{fig:unetcomparison}
\end{figure}

We compared the network performance to the polynomial activation, the inhibition model, and the attention model. Fig. \ref{fig:unetcomparison} shows the training and validation losses for the networks with different activation methods. The proposed network was trained much faster with a lower training loss than the other networks. Table \ref{tab:unetcomparison2} shows the performance and number of parameters of the networks. For object recognition, all variant activation methods provided a performance improvement. However, for denoising, only the inhibition model and proposed method provided a performance increase. The use of a learned but fixed polynomial activation or a heavier reliance on some parts with attention did not improve the MSE of the denoised images. The proposed method learned the interdependency features for denoising from data. U-net-AN provided the lowest MSE. Because U-net has more layers for upconversion than a network for a recognition task, the increase in complexity is higher than that observed for LeNet-AN for object recognition.

\begin{table}[h!]										
\centering										
\caption{Comparisons of Network Performance for U-net with MNIST}	
\label{tab:unetcomparison2}																		
\begin{tabular}{ll|cc|c}	
	\hline
	\multicolumn{2}{c|}{network} & \multicolumn{2}{c|}{\# of parameters} & MSE \\	\hline
	LeNet & baseline & 231041 &  (100.0\%)& 0.0144\\ \cline{2-5}
	& polynomial & 232641 & (100.7\%) & 0.0158\\ \cline{2-5}
	& inhibition & 231041 & (100.0\%)& 0.0144\\ \cline{2-5}
	& attention & 257985 & (111.7\%)& 0.0158\\ \hline
	LeNet-AN & & 365761 & (158.3\%)& 0.0133\\ \hline							
\end{tabular}
\end{table}

\subsection{VGG for Object Recognition}
\label{sec:vgg}

\begin{figure}[h]
	\centering
	\begin{minipage}{0.4\linewidth}
		\centering

		{\includegraphics[width=\linewidth]{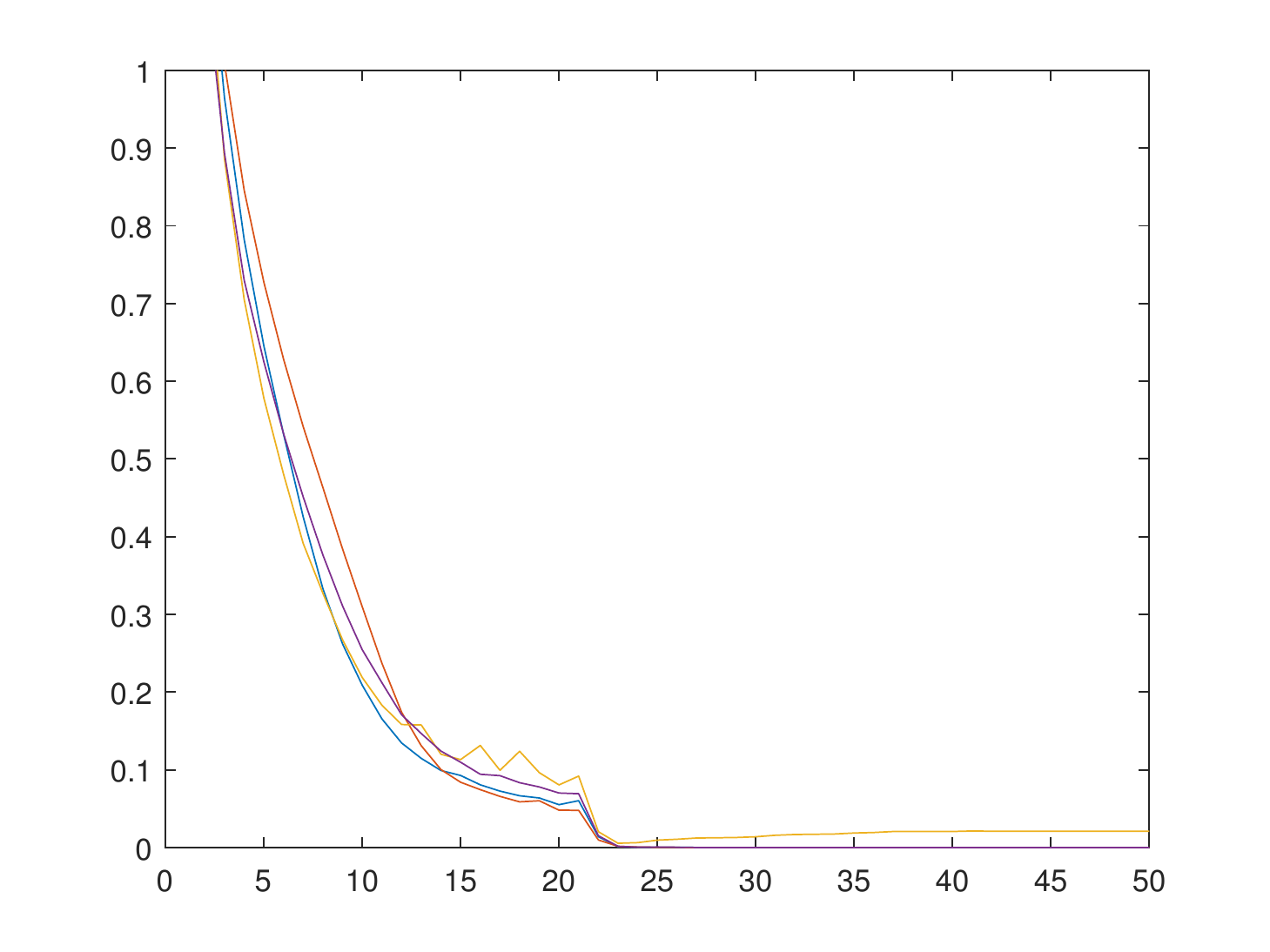}}%

		{\footnotesize (a)}
	\end{minipage}%
	\begin{minipage}{0.4\linewidth}
		\centering
		{\includegraphics[width=\linewidth]{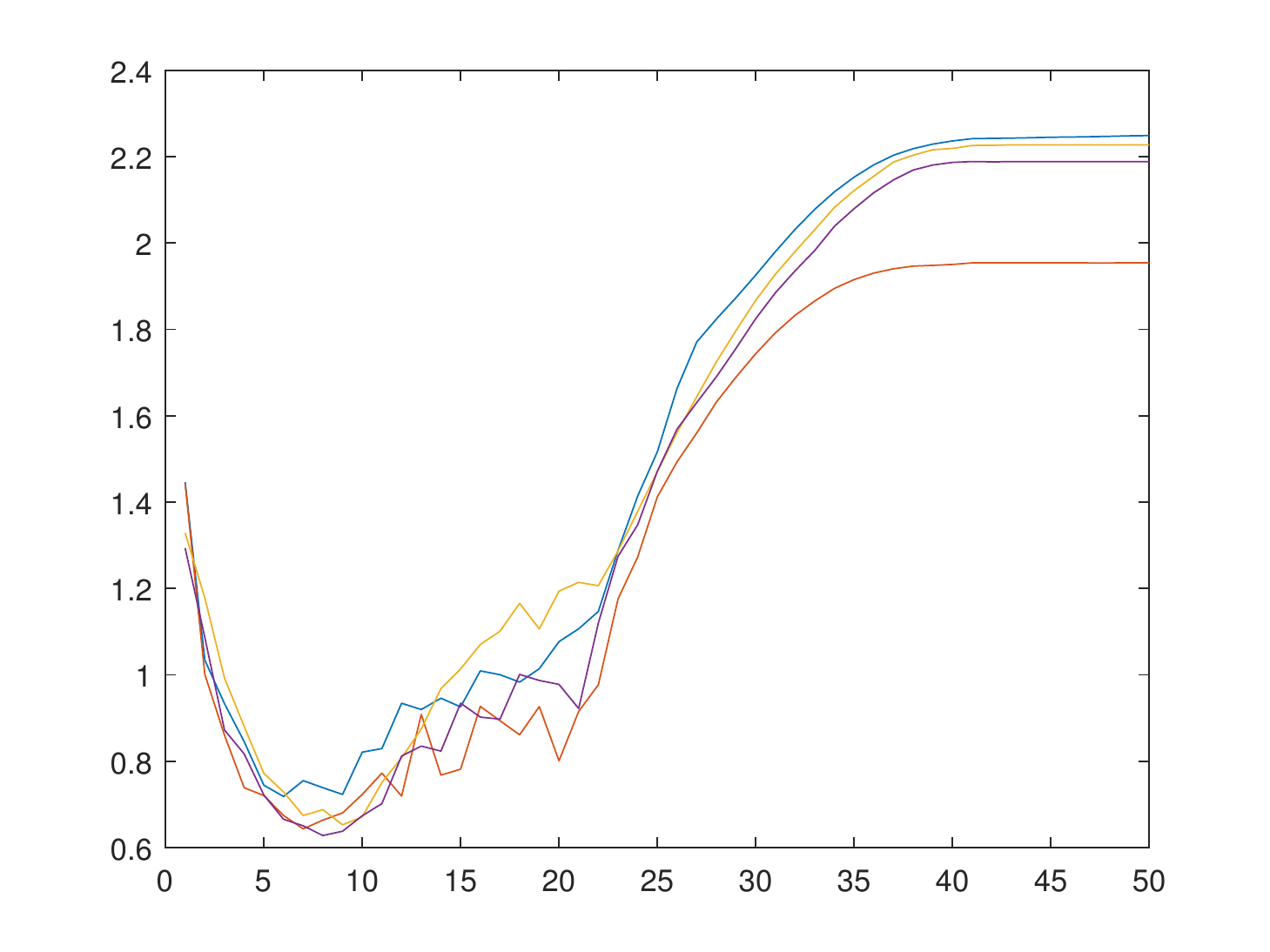}}%
		
		{\footnotesize (b)}
	\end{minipage}%
	
	\caption{Training of proposed VGG with CIFAR-10.
		(a) training loss and
		(b) validation loss.
		blue: baseline VGG;
		red: baseline VGGns;
		yellow: VGG-AN;
		purple: VGGns-AN.
		}
	\label{fig:vggtraining}
\end{figure}

In this section, we consider designing a transferred deeper network for object recognition. We prepared a deep network based on VGG16 \cite{krizhevsky2012imagenet} by adding activation networks. The VGG has cascades of processing blocks that consist of multiple convolutional layers and a pooling layer. We added an activation network after the first convolutional layer of each processing block. The first convolutional layer takes the output of a pooling layer and effectively uses the features with larger supports. By adding an activation network, we can impose how to activate the features based on the features with larger supports. The VGG16-based network with activation networks is denoted as VGG16-AN. For comparison, the baseline VGG16 with the ReLU activation function is also prepared. The networks are trained with the CIFAR10 dataset for object recognition. Fig. \ref{fig:vggtraining} shows the training and validation losses for the baseline VGG16 and VGG16-AN. VGG16-AN was trained much faster, with lower training and validation losses. 

Table \ref{tab:vggcomparison} shows comparisons of the network performance and complexity. VGG16-AN provides a higher accuracy than the baseline VGG16. The computational complexity of VGG16-AN is 120.5\% of that of the baseline VGG16. We prepared another network on VGG16 with less computational complexity. All processing blocks consist of two convolutional layers and a pooling layer. The number of nodes in each layer is reduced by half. The narrower and shallower networks are denoted as VGGns and VGGns-AN. Fig. \ref{fig:vggtraining} shows the training and validation losses. The network performance and complexity are also shown in Table \ref{tab:vggcomparison}. The computational complexity of VGGns-AN is only 21.7\% of that of the baseline VGG. However,  VGGns-AN outperforms both the VGGns and baseline VGG16. With activation networks to exploit the interdependency on features, the network with shallower layers and smaller nodes outperforms the much deeper network.

\begin{table}[h!]										
\centering										
\caption{Comparisons of Network Performance for VGG16 and CNN with CIFAR-10}	
\label{tab:vggcomparison}																		
\begin{tabular}{ll|cc|c}	
	\hline
	\multicolumn{2}{c|}{network} & \multicolumn{2}{c|}{\# of parameters} & test accuracy \\	\hline
	VGG16 & baseline & 14978250& (100.0\%) & 0.8097\%\\ \hline
	VGG16-AN & & 18046998 & (120.5\%) & 0.8360\%\\ \hline						
	VGGns & & 2409674& (16.1\%) & 0.8036\%\\ \hline
	VGGns-AN & & 3252694 & (21.7\%) & 0.8205\%\\ \hline							
\end{tabular}
\end{table}

\section{Conclusion}
\label{sec:conclusion}

Deep networks equipped with auxiliary activation networks are presented. The proposed method represents a comprehensive form of activation. Each pixel, node, and layer can be assigned with a different activation function. The activation of a feature depends on other features with the dependency learned from data. In our experiments, providing an adaptive activation that exploits the dependency of features with activation networks improved the network performance significantly at a modest increase in the computational complexity. The proposed method can be used to improve the network performance as an alternative to increasing the number of nodes and layers.

\bibliographystyle{IEEEtran}
\bibliography{IEEEabrv,./ref.bib}

\end{document}